\documentclass[10pt,twocolumn,letterpaper]{article}

\usepackage{cvpr}              

\usepackage[dvipsnames]{xcolor}

\usepackage{makecell}

\definecolor{cvprblue}{rgb}{0.21,0.49,0.74}
\usepackage[pagebackref,breaklinks,colorlinks,citecolor=cvprblue]{hyperref}

\def\confYear{2024}

\title{OpenSight: A Simple Open-Vocabulary Framework \\for LiDAR-Based Object Detection}

\author{Hu Zhang\textsuperscript{1}\thanks{This work is partially done during internship at Alibaba.} \quad
Jianhua Xu\textsuperscript{2} \quad
Tao Tang\textsuperscript{2} \quad
Haiyang Sun\textsuperscript{4} \\
Xin Yu\textsuperscript{1} \quad
Zi Huang\textsuperscript{1}\thanks{Denotes the corresponding author.} \quad
Kaicheng Yu\textsuperscript{3} \\
\textsuperscript{1}University of Queensland, Australia \\
\textsuperscript{2}DAMO Academy, Alibaba, China \\
\textsuperscript{3}Westlake University, \textsuperscript{4}Li Auto \\
{\tt\small \{hu.zhang, helen.huang\}@uq.edu.au}
}

\begin{document}
\maketitle

\begin{abstract}
    Traditional LiDAR-based object detection research primarily focuses on closed-set scenarios, which falls short in complex real-world applications. Directly transferring existing 2D open-vocabulary models with some known LiDAR classes for open-vocabulary ability, however, tends to suffer from over-fitting problems: The obtained model will detect the known objects, even presented with a novel category. In this paper, we propose OpenSight, a more advanced 2D-3D modeling framework for LiDAR-based open-vocabulary detection. OpenSight utilizes 2D-3D geometric priors for the initial discernment and localization of generic objects, followed by a more specific semantic interpretation of the detected objects. The process begins by generating 2D boxes for generic objects from the accompanying camera images of LiDAR. These 2D boxes, together with LiDAR points, are then lifted back into the LiDAR space to estimate corresponding 3D boxes. For better generic object perception, our framework integrates both temporal and spatial-aware constraints. Temporal awareness correlates the predicted 3D boxes across consecutive timestamps, recalibrating the missed or inaccurate boxes. The spatial awareness randomly places some ``precisely'' estimated 3D boxes at varying distances, increasing the visibility of generic objects. To interpret the specific semantics of detected objects, we develop a cross-modal alignment and fusion module to first align 3D features with 2D image embeddings and then fuse the aligned 3D-2D features for semantic decoding. Our experiments indicate that our method establishes state-of-the-art open-vocabulary performance on widely used 3D detection benchmarks and effectively identifies objects for new categories of interest.
\end{abstract}

\section{Introduction}
\label{sec:intro}

The growing prevalence of autonomous driving has significantly increased research interest in various 3D perception tasks~\cite{guo2020deep, yurtsever2020survey, douillard2011segmentation, simon2019complexer}, among which LiDAR-based object detection stands as a crucial component. Despite the advancements in this field, most existing research focuses on closed-set settings~\cite{qi2017pointnet++, lang2019pointpillars, liang2022bevfusion, zhou2018voxelnet, yan2018second}, where the obtained models are limited to predicting a set of predefined categories. This constraint makes these models insufficient for handling complex real-world scenarios, where diverse and previously unseen object categories may be encountered. Consequently, there is a growing demand to develop models capable of detecting novel categories of interest.


\begin{figure}
  \centering
  \begin{subfigure}[b]{0.27\textwidth}
      \includegraphics[width=\textwidth]{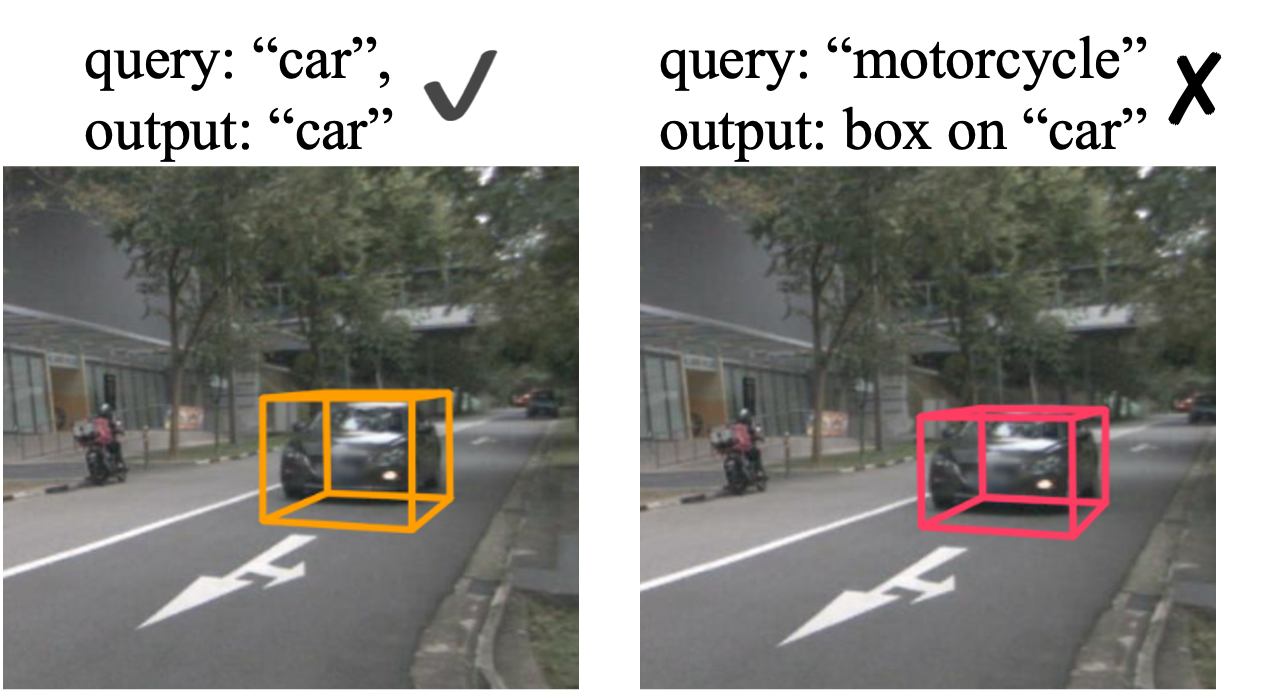}
      \caption{}
  \end{subfigure}
  \begin{subfigure}[b]{0.20\textwidth}
      \includegraphics[width=\textwidth]{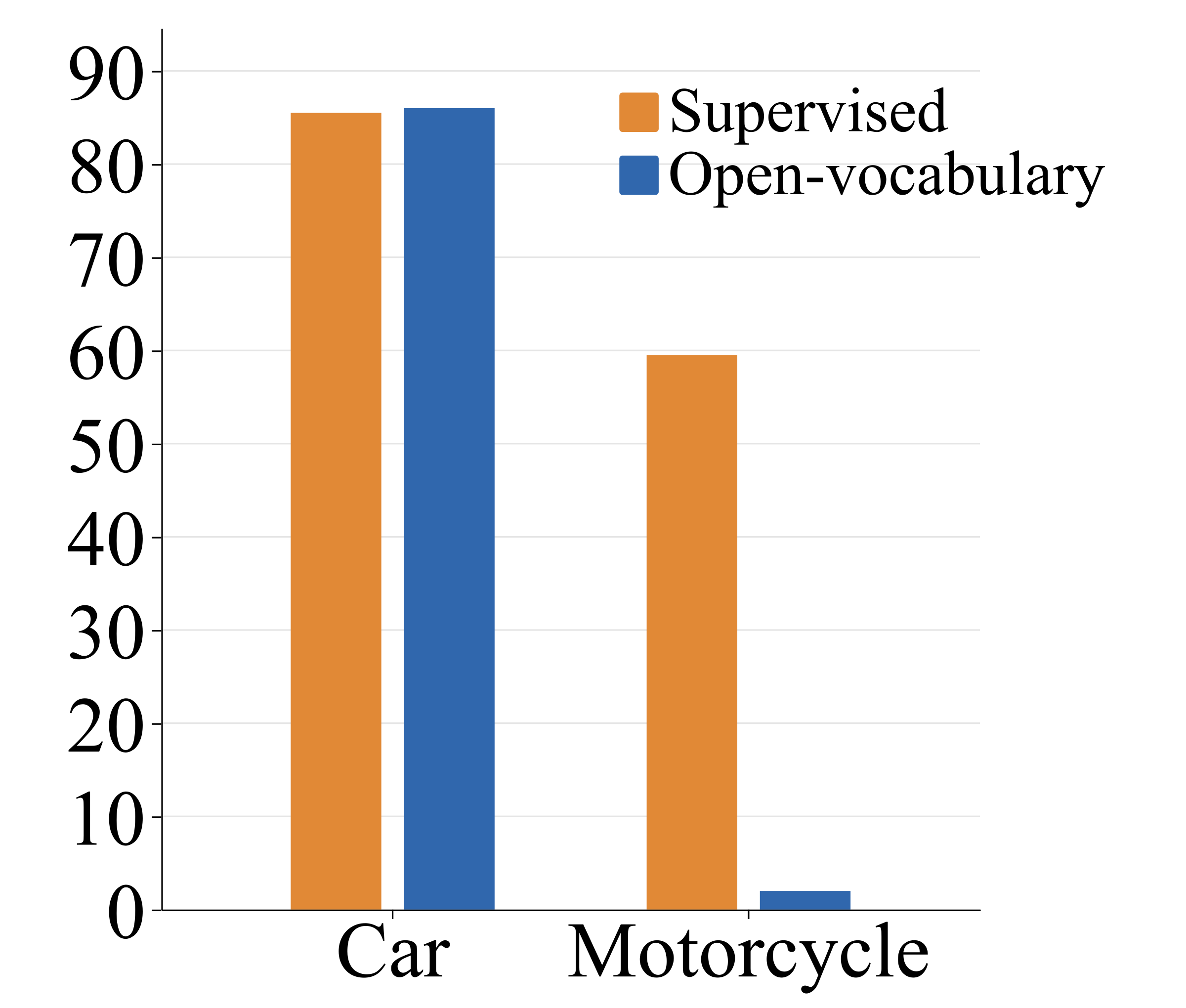}
      \vspace{-1.8em}
      \caption{}
  \end{subfigure}
  \vspace{-2em}
  \caption{(a) A model trained on known class ``car'' correctly detects the target object (\textcolor{yellow}{yellow}) with query ``car''. However, with a new class query ``motorcycle'', the model incorrectly detects the object (\textcolor{red}{red}).
  (b) The average precision of known ``car'' and novel ``motorcycle''. The degraded performance in ``motorcycle'' (\textcolor{blue}{blue}) indicates the limited generalization from the known classes.}
  \label{fig: the boxes of novel classes have been suppressed.}
  \vspace{-2em}
\end{figure}
Considering the limited availability of annotated categories in existing LiDAR-based datasets~\cite{Geiger2012CVPR, nuscenes, sun2020scalability}, adapting established large-scale visual-language models (LVLs), e.g., CLIP~\cite{radford2021learning} for open-vocabulary LiDAR detection appears promising. One paradigm is to co-embed known class object features with the image/text embeddings from LVLs, expecting alignment generalization to the novel classes. While effective in 2D image-based open-vocabulary detection, this paradigm exhibits significant overfitting in LiDAR applications, as models persist in detecting objects in known classes even when faced with a new category.

We consider some known classes, \emph{e.g.}, ``car'', and follow the paradigm above to illustrate this phenomenon. When given a novel category ``motorcycle'', the obtained model erroneously detects the known object ``car'', as depicted in Figure~\ref{fig: the boxes of novel classes have been suppressed.} (a). An analysis of average precision (AP) further reveals the diminished efficacy in novel classes, as shown in Figure~\ref{fig: the boxes of novel classes have been suppressed.} (b). These results underscore the limitation of such a paradigm, emphasizing the necessity for a tailored approach for LiDAR-based open-vocabulary problems.




In this paper, we present ``OpenSight'', a more advanced 2D-3D modeling approach for LiDAR-based open-vocabulary detection. OpenSight primarily relies on 2D and 3D geometry priors to first acquire the concept and localization of generic objects. Then, it interprets the semantics of detected objects by co-embedding LiDAR features with 2D LVLs. This framework addresses all potentially novel objects explicitly and circumvents the reliance on predefined known classes, thereby avoiding the overfitting problem.




Specifically, we consider both temporal-aware and spatial-aware constraints to acquire the concept of generic objects and their localization. Initially, leveraging the 2D detection model and the accompanying RGB images of LiDAR, we extract 2D geometric priors for generic objects, that is, 2D bounding boxes across successive frames. These boxes are then lifted back into 3D space with LiDAR points, resulting in a collection of 3D frustums. Point clustering is then performed in each frustum to estimate the 3D bounding boxes of objects. Temporally, we correlate predicted 3D bounding boxes across frames and project the predicted 3D boxes from previous or subsequent frames onto the current frame. The missed or inaccurate objects in the current frame are then identified through overlapping comparison. This constraint boosts the continuity and consistency of generic objects over time. Spatially, we first employ large language models (LLMs) to derive size priors for general objects and then maintain 3D boxes that closely align with these size priors. The maintained boxes are randomly selected and placed at varying distances within the current frame, with the LiDAR points inside these boxes being sampled by their distance. This spatial awareness notably increases the visibility of objects, particularly those located further from the sensor. By integrating the temporal and spatial constraints, the ability of the obtained model to recognize and localize generic objects is significantly augmented.

To decode the specific semantics of detected boxes in LiDAR space, we correlate objects across both LiDAR and image domains. Initially, by back-projecting 3D LiDAR bounding boxes onto the image plane, we identify the most closely matching 2D boxes from the grounding DINO. We then aim to minimize the feature discrepancy between the 3D LiDAR objects and the 2D matching box features. Subsequently, we fuse the optimized 3D-2D features and align the unified features with textual category embeddings from grounding DINO to decode semantics. This dual 2D-3D feature approach leverages salient patterns in both 2D images and 3D geometry, significantly enhancing the robustness of semantic decoding. Additionally, we integrate size priors from LLMs before, recalibrating predictions that diverge from the average box size, thus enhancing the accuracy of our predictive model.





In the experiments, we first establish a variety of baselines for comparison. Then, we test our method on the widely-used 3D detection benchmarks, demonstrating that our approach achieves state-of-the-art open-vocabulary performance and effectively identifies objects for new categories of interest. 


\section{Related Works}
We review previous works on object detection, encompassing closed-set 3D object detection, open-vocabulary 2D object detection, and zero-shot learning in 3D data.

\noindent \textbf{Closed-set 3D Object Detection} 
Extensive research has been proposed on LiDAR-based 3D object detection in autonomous driving. One group of works performs on LiDAR-only data~\cite{shi2020pv, zhou2020end}, either by extracting features from raw point clouds to predict 3D bounding boxes~\cite{qi2017pointnet++, qi2017pointnet, qi2018frustum, shi2019pointrcnn, yang20203dssd, li2021lidar} or converting the irregular LiDAR point clouds into a regular feature space~\cite{zhou2018voxelnet, yan2018second, lang2019pointpillars, wang2020pillar, yin2021center, fan2021rangedet, sun2021rsn}.
Another group focuses on both LiDAR and camera modalities.
Some works emphasize point-wise fusion~\cite{huang2020epnet, sindagi2019mvx, wang2021pointaugmenting, yin2021multimodal}, projecting raw LiDAR points to retrieve corresponding image features and concatenating them together.
In contrast, others concentrate on feature-level fusion~\cite{yoo20203d, liang2018deep, li2022deepfusion}, extracting features or proposals based on LiDAR points and querying the associated image features. This line of methods has been proven to be more efficient. For example, TransFusion~\cite{bai2022transfusion} utilizes proposals generated by LiDAR features to query the image feature for further refinement. BEVFusion~\cite{liang2022bevfusion, liu2022bevfusion} projects the image features into BEV space by using lift-splat-shoot (LSS) and concatenates them with the LiDAR features. More recently, LoGoNet~\cite{li2023logonet} constructs Local-to-Global fusion network to perform LiDAR-camera fusion at both local and global levels. However, all these approaches target the closed-set setting, restricting the range of detected object categories. 
\begin{figure*}
  \centering
  \includegraphics[width=1.\textwidth]{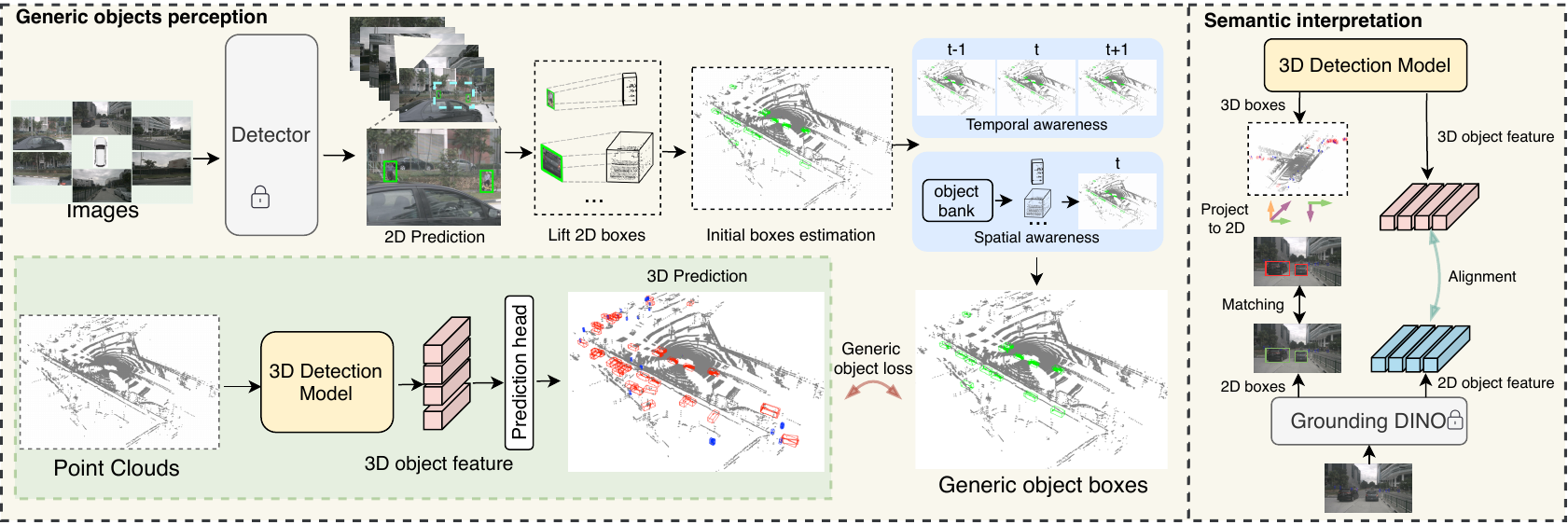}
  \caption{An overview of our OpenSight framework. In generic object perception, 2D bounding boxes are initially lifted back into 3D space (initial boxes estimation). This is followed by integrated temporal and spatial awareness in the estimated 3D boxes to retrieve the missed objects or increase their visibility. Subsequently, we utilize these enhanced estimates to compute the generic object loss and optimize the 3D detection model. In specific semantic interpretation, we project the generated 3D boxes of the optimized 3D model into the image plane to get projected 2D boxes. These projections are then matched with the nearest 2D boxes identified via grounding DINO, facilitating the alignment of 3D LiDAR object features with corresponding 2D image features.}
  \label{fig: overall framework.}
\end{figure*}

\noindent \textbf{Open-vocabulary 2D Object Detection}
The advent of large-scale pre-trained models, which exhibit good text-vision alignment and zero-shot transferability, has spurred researchers to investigate the open-vocabulary problem in object detection tasks. A variety of approaches have been proposed with 2D RGB images: OVR-CNN~\cite{zareian2021open} initially pre-trains a Faster R-CNN detector~\cite{ren2015faster} by using image-caption pairs and text extractor, e.g., BERT~\cite{devlin2018bert}, then fine-tunes it on downstream detection datasets. ViLD~\cite{gu2022openvocabulary} distills the knowledge from a CLIP-based~\cite{radford2021learning} teacher to a student detector for generalized text and image embedding alignment. Detic~\cite{zhou2022detecting} harnesses extensive image classification data to train the classifier within a detector while localizing objects. Similar to OVR-CNN, OWL~\cite{minderer2022simple} first pre-trains a transformer encoder by contrastively aligning image-text pairs, then transfers it to detection tasks by attaching light-weighted object classification and localization heads. Following the DETR-based structure, OV-DETR~\cite{zang2022open} employs CLIP-encoded text and image embeddings as queries for obtaining category-specific boxes. In contrast to the aforementioned methods, some alternatives have emerged to directly learn open-vocabulary capabilities on detection data. For example, GLIP~\cite{li2022grounded} unifies the object detection and phrase grounding tasks, allowing the combined use of both detection and grounding datasets for aligning phrase semantics and regional images.
DetCLIP~\cite{yao2022detclip} extracts the general concepts from multiple data sources and then resorts to a description-enriched concept dictionary for enhancement. Recently, Ground DINO~\cite{liu2023grounding} adopts a similar detection-grounding training strategy as GLIP but proposes a tighter text-vision fusion solution based on transformer-based detector DINO~\cite{zhang2022dino}.

\noindent \textbf{Zero-shot Learning in 3D Data}
Exploration of open-vocabulary 3D object detection has been limited due to the challenges associated with acquiring large-scale point cloud-text pairs and the inherent complexity of 3D scenarios. In the confined indoor scenario, OV-3DET~\cite{lu2023open} addresses the open-vocabulary detection problem by first localizing 3D objects and then assigning labels for the objects with open-vocabulary 2D models. However, temporal and spatial constraints are overlooked. CoDA~\cite{cao2023coda} proposes to use some base classes to assist in detecting novel classes. As illustrated in Figure~\ref{fig: the boxes of novel classes have been suppressed.}, such a paradigm in LiDAR data easily overfits the base classes. Some other works~\cite{cen2021open, wong2020identifying, alliegro20223dos,cen2022open} tackle a related but distinct task of differentiating unknown data in testing from known data in training, known as the out-of-distribution (OOD) problem. Meanwhile, some studies~\cite{peng2022openscene, michele2021generative, liu2021language} attempt to transfer existing 2D large-scale pre-trained models for open-vocabulary scene segmentation, while others~\cite{zhang2022pointclip, cheraghian2019mitigating, cheraghian2020transductive, cheraghian2022zero,cheraghian2019zero} leverage the open-set ability of pre-trained models for the 3D classification tasks.


Different from the above works, in this paper, we investigate open-vocabulary ability in LiDAR-based 3D object detection in the outdoor scenario, representing the real-world scenario and a much more complex scene.

\section{OpenSight: Open-Vocabulary LiDAR-based Object Detection}
To elaborate on the LiDAR-based open-vocabulary object detection, we first briefly introduce the necessary notations for ease of understanding. Following this, we delve into the details of generic object perception in Section~\ref{subsection: generic object perception}. Subsequently, we present the process of specific semantic interpretation in Section~\ref{subsection: specific semantics interpretation}. The overall framework of OpenSight is presented in Figure~\ref{fig: overall framework.}.


\subsection{Notation and Preliminaries}
The LiDAR dataset is denoted as $\mathcal{D} = \{P_i, \{y_k \}_{k=1}^K\}_{j=1}^{|\mathcal{D}|}$, where $P_i$ corresponds to the LiDAR point clouds in $i$-th frame, basically composed of $(x_i, y_i, z_i)$ coordinates. $y_k$ represents the combination of each 3D bounding box $b_k$ and its semantic label $c_k$. In open-vocabulary scenarios, $P_i$ is visible while $y_k$ is not. Generally, in the context of autonomous driving, each LiDAR frame is accompanied by $m$ images captured from varying perspectives. We define the image dataset as $\mathcal{I} = \{(I_{ij})_{j=1}^m\}_{i=1}^{|\mathcal{D}|}$ here. Image $I_{ij}$ in $j$-th view and LiDAR points $P_i$ could be connected by the intrinsic matrix $\textbf{K}_j$ and extrinsic matrix $\textbf{E}_j$. For clarity and brevity in the rest of the paper, we utilize a single LiDAR frame as an example and abbreviate and $I_{ij}$ as $I_j$, respectively. $N$ denotes the number of 3D objects, with $f^{3D}$ and $f^{2D}$ denotes the extracted feature in 3D and 2D space.


\subsection{Generic Objects Perception}
\label{subsection: generic object perception}
Discerning and localizing generic objects are fundamental in LiDAR-based open-vocabulary detection when ground-truth boxes and labels are absent. The model is expected to perceive the existence of a generic object before assigning specific labels.
This section details how we leverage 2D and 3D geometry priors to enable this perception, including initial box estimation, and temporal-spatial awareness. 

\noindent{\textbf{Initial Box Estimation.}} In practical applications, each LiDAR frame is typically paired with $m$ images $\{I_j\}_{j=1}^{m}$, captured from various perspectives. Each image is processed through the pre-trained detection model $\mathcal{I}$, yielding a set of 2D bounding boxes $\{b_{jk}\}_{k=1}^{V_j}$, where $V_j$ represents the number of 2D objects identified in the $j$-th viewpoint. By integrating the data from all viewpoints, we compile the collective set of detected boxes $\{(b_{jk}^{2D})_{k=1}^{V_j}\}_{j=1}^{m}$.

With the obtained 2D boxes of generic objects, we try to lift them back in the LiDAR space. Specifically, we project the LiDAR point $P_i$ into each image view by:
\begin{equation}
    \hat{P}_{ij} = \textbf{K}_j \cdot \textbf{E}_j \cdot P_i, \  j=1,...,m
\end{equation}
where $\textbf{E}_j$ and $\textbf{K}_j$ are extrinsic and intrinsic matrix, respectively. The points whose projection falls within each of the predicted 2D bounding boxes $b_{jk}^{2D}$ essentially constitute the 3D frustum, a conical region perceivable from a given perspective. We then apply the region growing algorithm~\cite{adams1994seeded} to isolate the cluster with the densest LiDAR points.

To estimate 3D bounding boxes from LiDAR point clusters, we first determine the extremal points along each axis by calculating the minimum and maximum coordinates. Take the X-axis for example, this is achieved by performing $x_{\min} = \min(\{x_i\}_{i=0}^{|N_c|})$ and $
x_{\max} = \max(\{x_i\}_{i=0}^{|N_c|})$.
These extremal points, identified similarly for the Y and Z axis, define the vertices of the bounding box. From these vertices, we construct the maximal bounding box $b_i^{3D}$, enclosing all points in the cluster. The 3D boxes for generic objects at timestamp $t$ are then finalized as $\mathcal{B}^t = \{b_{i}^{3D}\}_{i=0}^{N_t}$, by aggregating all boxes from $m$ perspectives.

\begin{figure}[t]
  \centering
      \includegraphics[width=0.45\textwidth]{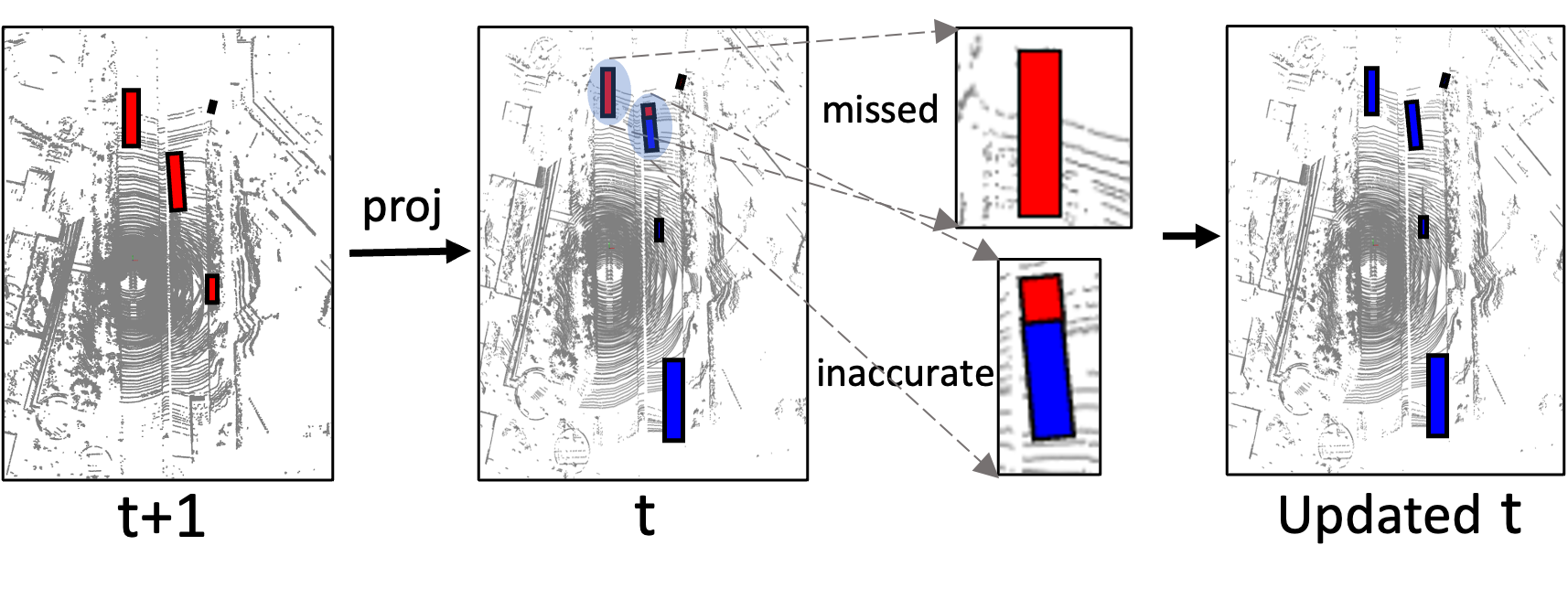}
  \vspace{-1em}
  \caption{Temporal awareness: Project boxes from frame $t+1$ to $t$. By comparing with projections, we refine predicted boxes in frame $t$ for missed or inaccurately estimated objects.}
  \label{fig: temporal.}
  \vspace{-2em}
\end{figure}
Considering that the majority of objects are lying on the ground and the occluded objects from one perspective, we adopt two filtering strategies to improve frustum construction. Firstly, to mitigate the impact of ground points on LiDAR point projection and clustering, we use RANSAC~\cite{fischler1981random} for ground plane estimation and exclude these points. 
Secondly, to remove the distortion from bounding box corners, largely caused by other overlapped objects, we apply the SAM method~\cite{kirillov2023segany}. By inputting the detected 2D boxes as prompts, we can obtain more accurate object contours.


\noindent \textbf{Temporal Awareness.}
For the LiDAR point cloud, point density decreases sharply as distance increases, leading to less reliable estimations for distant objects. However, in autonomous driving scenarios, objects that are distant in the current frame may move closer in subsequent frames. To exploit this, we leverage the temporal continuity between frames by correlating 3D bounding box predictions across successive timestamps. We take frames at timestamp $t$ and $t+1$ as an example. Specifically, given the predicted boxes $\mathcal{B}^t = \{b_i\}_{i=0}^{N_t}$ and $\mathcal{B}^{t+1} = \{{b_i\}_{i=0}^{N_{t+1}}}$, where $N_{t+1}$ and $N_t$ represent the number of boxes in $t+1$ and $t$, respectively, we project each 3D box $b_i$ in $\mathcal{B}^{t+1}$ back to frame $t$ using the transformation:
\begin{equation*}
\hat{b}_i
= \textbf{E}_{\text{lidar} \leftarrow \text{ego}_{t}} \cdot
\textbf{E}_{\text{ego}_{t} \leftarrow \text{glb}} \cdot
\textbf{E}_{\text{glb} \leftarrow \text{ego}_{t+1}} \cdot
\textbf{E}_{\text{ego}_{t+1} \leftarrow \text{lidar}} \cdot
b_i,
\end{equation*}
where $\textbf{E}_{\text{lidar} \leftarrow \text{ego}_{t}}$ and 
$\textbf{E}_{\text{ego}_{t} \leftarrow \text{glb}}$ are the extrinsic matrices for the ego to LiDAR and global to ego transformations at timestamp $t$, and $\textbf{E}_{\text{glb} \leftarrow \text{ego}_{t+1}}$,
$\textbf{E}_{\text{ego}_{t+1} \leftarrow \text{lidar}}$ denotes ego to global and LiDAR to ego at timestamp $t+1$, respectively.

With all boxes in $\mathcal{B}^{t+1}$ considered, we obtain their projections as $\hat{\mathcal{B}}^{t+1}$ and construct the overlap matrix $M \in \mathbb{R}^{N_{t+1}\times N_t}$ by calculating their Intersection over Union (IoU) with boxes in $\mathcal{B}^{t}$. We address two key scenarios: If a box $\hat{b}_i$ has no overlap with boxes in $\mathcal{B}^{t}$, its counterpart $b_i$ in timestamp $t+1$ is considered a missed detection in the current frame and is added to the prediction at timestamp $t$. Conversely, if box $\hat{b}_i$ overlaps with boxes in $\mathcal{B}^{t}$ and its distance exceeds a certain threshold, we perform Non-Maximum Suppression (NMS) and retain the union of the two boxes to address potentially incomplete estimations at greater distances (See Figure~\ref{fig: temporal.}). 

This process updates the predicted boxes of the current frame to $\mathcal{B}^t_{\rm update} = \mathcal{B}^t \bigcup \mathcal{B}^t_{\rm temporal}$, which combines the original and temporally adjusted boxes. In practice, for the current frame, we consider both its subsequent frame at timestamp $t+1$ and its precedent frame at timestamp $t-1$. Through such a constraint, we are likely to boost the continuity and consistency of generic objects over time.


\begin{figure}[t]
  \centering
      \includegraphics[width=0.48\textwidth]{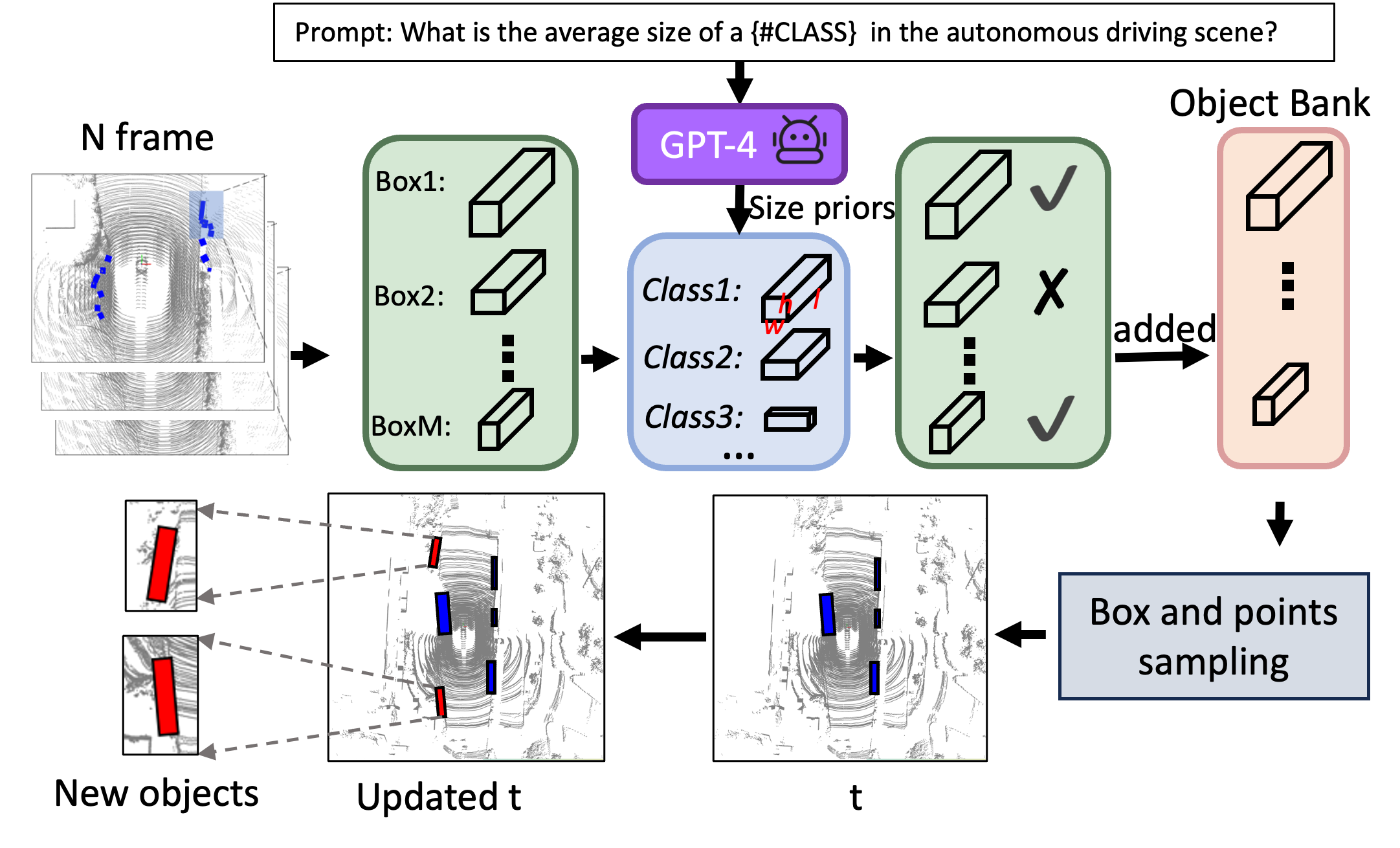}
  \vspace{-2em}
  \caption{Spatial awareness: Size priors are first generated from GPT-4 to select ``precisely'' estimated boxes and build the object bank. Randomly select objects and points from the objects bank and place them in the current frame $t$.}
  \label{fig: spatial.}
  \vspace{-2em}
\end{figure}
\noindent \textbf{Spatial Awareness.}
Considering the sparse object distribution in 3D LiDAR scans, where vast areas of the point cloud are often devoid of meaningful objects, we also focus on the systematic integration of ``precisely'' estimated objects in each frame to enhance the detectability and representation of generic objects.
To obtain these precise boxes from $\mathcal{B}^t$ in the absence of any ground truth, we utilize large language models (LLMs), such as GPT-4, for extracting useful size priors of objects. GPT-4, when prompted appropriately, can provide valuable insights into the typical sizes of objects within a specific class, \emph{e.g.}, the width, length, and height of a class. For instance, prompts like ``What is the average size of a \{\#CLASS\} in the autonomous driving scene?'' enable us to collect size statistics for common classes~(Details in the Appendix). Moreover, leveraging the ability of GPT-4 to process images, we also input images of the relevant objects to validate these size priors (See Figure~\ref{fig: spatial.}).

Utilizing the extracted size priors, we selectively retain 3D boxes in $\mathcal{B}^t$ that closely align with these priors. A box is reserved if its dimensions $w$, $l$, and $h$ fall within the specified range:
\begin{equation}
\label{equ: select box}
\begin{aligned}
    w &\in [(1-\tau)w_{prior}, (1+\tau)w_{prior}], \\
    \ell &\in [(1-\tau)\ell_{prior}, (1+\tau)\ell_{prior}], \\
    h &\in [(1-\tau)h_{prior}, (1+\tau)h_{prior}],
\end{aligned}
\end{equation}
where $\tau$ represents the error margin. $w_{prior}$, $l_{prior}$ and $h_{prior}$ are from LLMs-derived priors. This process results in an object bank $O = \{(b_i^{3D}, P_i)\}_{i=0}^{N_o}$, where $b_i^{3D}$ is the reserved box, $P_i$ comprises the LiDAR points within each box and $N_o$ is the total number of reserved boxes. 

For the current frame, we randomly select boxes from bank $O$ and integrate them at varying distances from the ego vehicle, ensuring diverse spatial distribution. The LiDAR points within each box are sampled according to their distance, with the sampling ratio defined as:
\begin{equation}
\label{equ: place object}
    r_{sampling} = \min(1, \frac{d-d_{new}}{d-d_{ori}}),
\end{equation}
where $d$ denotes the total range in the point cloud, $d_{ori}$ is the original distance of the object, $d_{new}$ is the distance post-placement. The selected boxes are denoted as $\mathcal{B}^t_{\rm spatial}$ and the collective set of boxes in the current frame is represented as $\mathcal{B}^t_{\rm final} =\mathcal{B}^t \bigcup \mathcal{B}^t_{\rm temporal} \bigcup \mathcal{B}^t_{\rm spatial}$.
This strategy aims to substantially enhance the visibility of generic objects in LiDAR data and the perception ability of the obtained model.



\noindent \textbf{Loss for Generic Object Perception.} With the obtained 3D boxes $\mathcal{B}^t_{\rm final} = \{b_{k}^{3D}\}_{k=1}^{V_{final}}$, we formulate the loss for generic object concept acquisition and localization. Utilizing a DETR-like structure, we extract features from LiDAR points and apply bipartite matching to determine the optimal permutation $\hat{\sigma}$, the loss is thus expressed as:
\begin{equation}
\vspace{-1em}
    \mathcal{L} = \sum_{k=1}^{N}[ \mathcal{L}_{\rm focal}(s_{\hat{\sigma}_i(k)}) + \mathcal{L}_{\rm L_1}(b_{\hat{\sigma}_i(k)}, b^{3D}_{k})],
\end{equation}
where $s_{\hat{\sigma}_i(k)}$ denotes whether it is an object or not, $\hat{\sigma}_i(k)$ is the element in $\hat{\sigma}$. $\mathcal{L}_{\rm focal}$ constrains the level of ``objectness'' of generated 3D boxes, $\mathcal{L}_{\rm box}$ considers the constraint on boxes. With this composite loss optimized, we obtain a model with the ability to perceive generic objects.


\subsection{Specific Semantics Interpretation}
\label{subsection: specific semantics interpretation}
To interpret the specific semantics of detected generic objects in LiDAR, we correlate objects across both LiDAR and image domains. We apply the model after generic object perception to a given LiDAR frame, producing boxes $\{b_i^{3D}\}_{i=0}^N$ and their corresponding features $\{f_i^{3D}\}_{i=0}^N$. Instead of using the image-level LVLs, \emph{e.g.}, CLIP, we opt for object-level LVLs, such as grounding DINO, which provides 2D boxes $\{b_{i=0}^{2D}\}_{i=0}^{N_{2D}}$, 2D object features $\{f_i^{2D}\}_{i=0}^{N_{2D}}$ and text embeddings $\{t_i\}_{i=0}^{N_{2D}}$ simultaneously.

For each detected 3D box $b_i^{3D}$, we project it to each view image with the corresponding intrinsic matrix $K_j$ and extrinsic matrix $E_j$, formulated as:
\begin{equation}
    \hat{b}_{ij}^{2D} = \textbf{K}_j \cdot \textbf{E}_j \cdot b_i^{3D}, \  j=1,...,m
\end{equation}
The $\hat{b}_{ij}^{2D}$ is further filtered and only those lying in the image are reserved.
For the reserved box $\hat{b}_{ij}^{2D}$, we compute the IoU with the 2D boxes $\{b_i^{2D}\}_{i=0}^{N_{2D}}$ from grounding DINO and find the most close one $b_j^{2D}$. The corresponding text embeddings $t_j^{2D}$ feature is retrieved. We then try to minimize the discrepancy between $f_i^{3D}$ and all 2D features that has the same text embeddings $t_j^{2D}$. Mathematically, it is expressed as: 
\begin{equation}
    \ell_{align} = -\frac{1}{N} \sum_{i=1}^{N} \log \frac{\sum_{j=0}^{N_i}e^{f_i^{3D} \cdot f_j^{2D}/ \delta}}{\sum_{k=0}^{N^{2D}} e^{f_i^{3D} \cdot f_k^{2D} / \delta}},
\end{equation}
where $N_i$ denote all 2D features that has the same text embeddings $t_j^{2D}$ for $b_i^{3D}$. Once the object-level alignment is finished, we can directly perform open-vocabulary LiDAR detection with the LiDAR input.

\noindent \textbf{2D-3D Integration.}
Considering that paired RGB images with LiDAR are always available in autonomous driving, we thus can perform 2D-3D feature integration. For each detected object, we have optimized 3D features $f_i^{3D}$ and identified the best matched 2D boxes feature $f_i^{2D}$, the similarity between 3D-2D features with the text embeddings in $\{t_i\}_{i=0}^{N_{2D}}$ are computed as follows:
\begin{equation}
\begin{aligned}
    s_j^{3D} &= \cos(f_i^{3D}, t_j), \\
    s_j^{2D} &= \cos(f_i^{2D}, t_j), \ j=0,...,N^{2D}
\end{aligned}
\end{equation}
We obtain the maximum value for 2D feature and 3D feature by $s^{3D} = \max_{j}\{s_j^{3D}\}_{j=0}^{N_{2D}}$,  $s^{2D} = \max_{j}\{s_j^{2D}\}_{j=0}^{N_{2D}}$, along with their semantic labels $c^{3D}$ and $c^{2D}$. The final label for object $i$ is determined by comparing $s_i^{2D}$ and $s_i^{3D}$, selecting the label associated with the larger value.

\noindent \textbf{Large Language Model (LLMs) Post-processing.}
The size priors from LLMs can be used to assist in recalibrating predictions. By matching the 3D IoU of predicted boxes with these size priors, we can determine the most matching prior and the corresponding sematic label. We then incorporate the predictions from size prior with results from 2D-3D integration by selecting label with the highest similarity.


\subsection{Inference}
we first extract 3D object features from the LiDAR point cloud and align them with text embeddings from novel categories. LLMs post-processing is also applied in assisting in the determination of the final labels. Additionally, when RGB images are available, we enhance our detection process by integrating 3D and 2D features, thereby improving the accuracy of our semantic label determination.

\section{Experiments}
\subsection{Experimental settings}
\label{sec: experimental settings}
\noindent \textbf{Dataset.}
We conduct our open-vocabulary experiments on nuScenes~\cite{nuscenes} dataset, a large-scale, outdoor dataset featuring multimodal data modalities. Each frame in the dataset contains data from one 32-beam LiDAR and six cameras, offering a comprehensive view of the surroundings. The dataset comprises 1000 scenes in total, with up to 1.4 million annotated 3D bounding boxes spanning 10 classes. We reserve all the classes as novel to evaluate the effectiveness of our proposed OpenSight. 

\noindent \textbf{Evaluation Metrics.}
For quantitative evaluation, we initially employ Average Precision (AP) on each class and mean Average Precision (mAP) across all novel classes. Considering LiDAR point density variation with distance, we introduce distance-based mAP metrics: mAP-near, mAP-midrange, and mAP-far. Similarly, acknowledging the size disparities among objects in different classes (e.g., `bus' vs. `traffic\_cone'), we categorize classes into `large', `medium', and `small' and propose mAP-large, mAP-medium, and mAP-small. More details are in the Appendix.

\noindent \textbf{Implementation.}
We use VoxelNet~\cite{zhou2018voxelnet} as our LiDAR backbone and voxelize the LiDAR point cloud by the default settings in~\cite{yin2021center, bai2022transfusion}. For designed novel classes, we eliminate the usage of ground truth labels in training. We use the ``bert-base-uncased'' encoder in grounding DINO for text embedding extraction. 
For convenience, we also apply grounding DINO to extract 2D boxes. The accompanying images with a high resolution of 900 $\times$ 1600 are directly fed into grounding DINO for accurate 2D box generation and corresponding object-level features generation. The number of object queries is set to 200. We adopt the AdamW optimizer and jointly optimize the 3D loss in known classes and the 2D loss in novel classes for 20 epochs. The learning rate and weight decay are set as 1.0e-4 and 0.01, respectively. The code is implemented in PyTorch with the open-sourced MMDetection3D~\cite{mmdet3d2020}.
\begin{table*}[t]
\footnotesize
  \centering
  \setlength{\tabcolsep}{6pt}
  \begin{tabular}{l|cc|cccccccccc}
    \toprule
    Method   & mAP & NDS &\scriptsize Car &\scriptsize Truck &\scriptsize C.V. &\scriptsize Bus &\scriptsize Trailer &\scriptsize Barrier &\scriptsize Bicycle &\scriptsize Traffic\_cone &\scriptsize Motorcycle  &\scriptsize Pedestrian\\
    \midrule
    Supervised~(box+label) & 57.3 &61.7 & 86.0 & 38.8 & 18.3 & 63.3 & 34.2 &70.5 &46.7 &70.2 &59.5 &85.2\\
    Supervised~(box only) & 34.2 &39.0 & 53.2 & 20.1 & 7.4 & 20.5 & 0.7 &50.1 &37.8 &52.6 &39.1 &60.8\\
    \midrule
    Detic-3D~\cite{zhou2022detecting} &5.8 &12.7 &15.1 &2.9 & 2.1 & 4.2 &0.0 & 0.1 & 6.2 & 0.8 & 6.1 & 21.1\\
    OWL-ViT-3D~\cite{minderer2022simple} &5.4 &12.4 &14.2 &2.5 & 1.8 & 4.0 &0.0 & 0.1 & 5.2 & 1.1 & 5.5 & 19.4\\
    \midrule
    OV-PointCLIP~\cite{zhang2022pointclip} &0.1 &5.0 &0.0 &0.0 &0.0 &0.0 &0.0 &0.0 &0.7 &0.2 &0.0 &0.1 \\
    OV-PointCLIPv2~\cite{Zhu2022PointCLIPV2} &2.8 &10.9 &0.0 &2.7 &0.0 &0.0 &0.0 &22.0 &0.0 &0.2 &0.0 &3.2\\
    CLIP-3D~\cite{radford2021learning} &7.2 &14.2 & 9.6 & 2.8 & 2.1 &2.6 &0.0 & 0.1 &11.5 & 14.4 &9.8 & 18.9\\
    OV-3DET~\cite{lu2023open} &5.7 &12.0 &16.1 &2.9 &0.3 &1.3 &0.0 & 0.3 &1.8 &11.0 &2.0 &21.1\\
    CoDA~\cite{cao2023coda} &10.3 &16.1 &85.5 &1.9 &0.0 &2.0 &0.0 &0.0 &3.9 &0.0 & 5.9 &3.5 \\
    \midrule
    OpenSight~(Ours) & 23.5 &24.0 & 32.1 & 11.6& 8.7 & 5.1 & 0.8 & 42.2 & 26.1 & 25.6 & 30.4 & 52.5\\
    \bottomrule
  \end{tabular}
  \vspace{-1em}
  \caption{Results on NuScenes dataset. All classes are reserved as novel classes. AP of each class and mAP across all classes are shown. ``Car'' in CoDA is the known class and is fully supervised. (`C.V' is denoted as `construction\_vehicle' here.)}
  \label{tab: overall results}
  \vspace{-1em}
\end{table*}
\subsection{Baselines}
As there are no direct baselines addressing open-vocabulary LiDAR-based 3D detection, we primarily construct the baselines based on well-known works~\cite{zhou2022detecting, Zhu2022PointCLIPV2, radford2021learning, cao2023coda, lu2023open} 
and compare with OpenSight. Specifically, they are:

\noindent \textbf{Fully-supervised training.} Initially, we perform fully supervised training across all categories, establishing an upper bound for the open-vocabulary detection performance. We consider using both ground-truth boxes and labels (`Supervised~(box+label)') or just using ground-truth boxes (`Supervised~(box only)'). 


\noindent \textbf{2D open-vocabulary methods.} Detic~\cite{zhou2022detecting} is a pioneer work in the realm of 2D open-vocabulary object detection tasks and can detect 21K classes. It is thus desirable for our study here. OWL-ViT~\cite{minderer2022simple} decodes the object labels and boxes directly from the pre-trained features and shows good robustness in real-world applications. With the 2D detection results of either Detic or OWL-ViT, we adopt the same procedure in Section~\ref{subsection: generic object perception} to obtain the corresponding 3D bounding boxes in LiDAR space. These two baselines are denoted as Detic-3D and OWL-ViT-3D.

\begin{table}
\footnotesize
  \centering
  \begin{tabular}{l|ccc}
    \toprule
    Method & mAP-near & mAP-midrange &mAP-far\\
    \midrule
    Detic-3D~\cite{zhou2022detecting} &18.5 &9.3 & 5.8 \\
    OWL-ViT-3D~\cite{minderer2022simple} &17.3 &8.2 &5.4\\
    OV-PointCLIP~\cite{zhang2022pointclip} &0.2 &0.1 &0.1 \\
    OV-PointCLIPv2~\cite{Zhu2022PointCLIPV2} &4.5 &3.6 &2.8 \\
    CLIP-3D~\cite{radford2021learning} &18.8 &11.2 &7.2\\
    OV-3DET~\cite{lu2023open} &16.1 &9.9 &5.7\\
    CoDA~\cite{cao2023coda} &18.2 &16.6 &10.3\\
    OpenSight (Ours) &28.0 &26.5 & 23.5\\
    \bottomrule
  \end{tabular}
  \vspace{-1em}
  \caption{Performance \emph{w.r.t} mAP-near, mAP-medium, mAP-far.}
   \label{tab: evaluation of distance}
  \vspace{-1em}
\end{table}
\noindent \textbf{3D open-vocabulary methods.} We then consider the latest open-vocabulary point cloud classification methods and a variety of indoor open-vocabulary works to evaluate their performance in our tasks here. Specifically, we utilize a fully-supervised method to generate pseudo boxes and employ PointCLIP~\cite{zhang2022pointclip}, PointCLIPv2~\cite{Zhu2022PointCLIPV2}, denoted as ``OV-PointCLIP'', ``OV-PointCLIPv2''. We also combine our predicted boxes and CLIP~\cite{radford2021learning} for this task (``CLIP-3D''). Meanwhile, we adapt the latest works OV-3DET~\cite{lu2023open} and CoDA~\cite{cao2023coda} in indoor settings to our task here and demonstrate the effectiveness of our proposed methods. 



\begin{table}
\footnotesize
  \centering
  \begin{tabular}{l|ccc}
    \toprule
    Method & mAP-large &mAP-medium &mAP-small\\
    \midrule
    Detic-3D~\cite{zhou2022detecting} &5.3 &4.1 & 11.0\\
    OWL-ViT-3D~\cite{minderer2022simple} &4.9 &3.7 &10.3\\
    OV-PointCLIP~\cite{zhang2022pointclip} &0.0 &0.3 & 0.2\\
    OV-PointCLIPv2~\cite{Zhu2022PointCLIPV2} &0.6 &7.3 &1.6\\
    CLIP-3D~\cite{radford2021learning} &3.4 &7.1 & 16.7 \\
    OV-3DET~\cite{lu2023open} &4.1 &1.4 &16.0 \\
    CoDA~\cite{cao2023coda} &17.9 &4.9 &1.8 \\
    OpenSight (Ours) &11.5 &32.9 & 39.0\\
    \bottomrule
  \end{tabular}
    \vspace{-1em}
    \caption{Performance \emph{w.r.t} mAP-large, mAP-medium, mAP-small.}
  \label{tab: evaluation of object sizes}
  \vspace{-1em}
\end{table}

\subsection{Main results}
In this section, we discuss the performance of our model on novel classes, as summarized in Table~\ref{tab: overall results}. Despite the lack of 3D ground truth supervision, our framework demonstrates strong overall open-vocabulary ability across classes. Notably, it achieves exceptional performance in some classes, for example, 52.5 AP in `pedestrian' and 42.2 AP in `Barrier'. Unfortunately, we also notice the limited performance in `Trailer', with 0.8 AP. To analyze this, we compare with `Supervised~(box only)', where groundtruth boxes are used and only 0.7 AP is achieved in `Trailer', we thus attribute the reason to label alignment with grounding DINO rather than box inaccuracy. Grounding DINO offers limited help in fine-grained classes. Our model also shows superior performance when compared with the baselines.  Specifically, we observe that PointCLIP is very limited in detecting novel objects and CLIP-3D achieves the best overall performance with our detected boxes. CoDA, while effective in known classes like `car', falls short in novel class adaptation, averaging only 1.9\% accuracy across novel categories.

Our more detailed evaluations, using distance-based metrics and size-based metrics, are presented in Tables~\ref{tab: evaluation of distance} and~\ref{tab: evaluation of object sizes}, respectively. The evaluation in Table~\ref{tab: evaluation of distance} reveals a trend where detection accuracy is higher for objects closer to the sensor. This is attributed to their denser point clouds and more defined geometric features. Conversely, detection accuracy decreases with increasing distance, likely due to sparser data and less distinct object outlines. Regarding size-based performance (Table~\ref{tab: evaluation of object sizes}), our model shows superior results for medium and small-sized objects. Large objects present lower detection rates, mainly due to their incomplete LiDAR scans. Additionally, the fine-grained categorization of large vehicles, such as `Trailer' and `Construction\_vehicle', complicates their accurate detection.


\subsection{Qualitative results on novel classes}
\begin{figure*}[t]
  \centering  \includegraphics[width=0.85\textwidth]{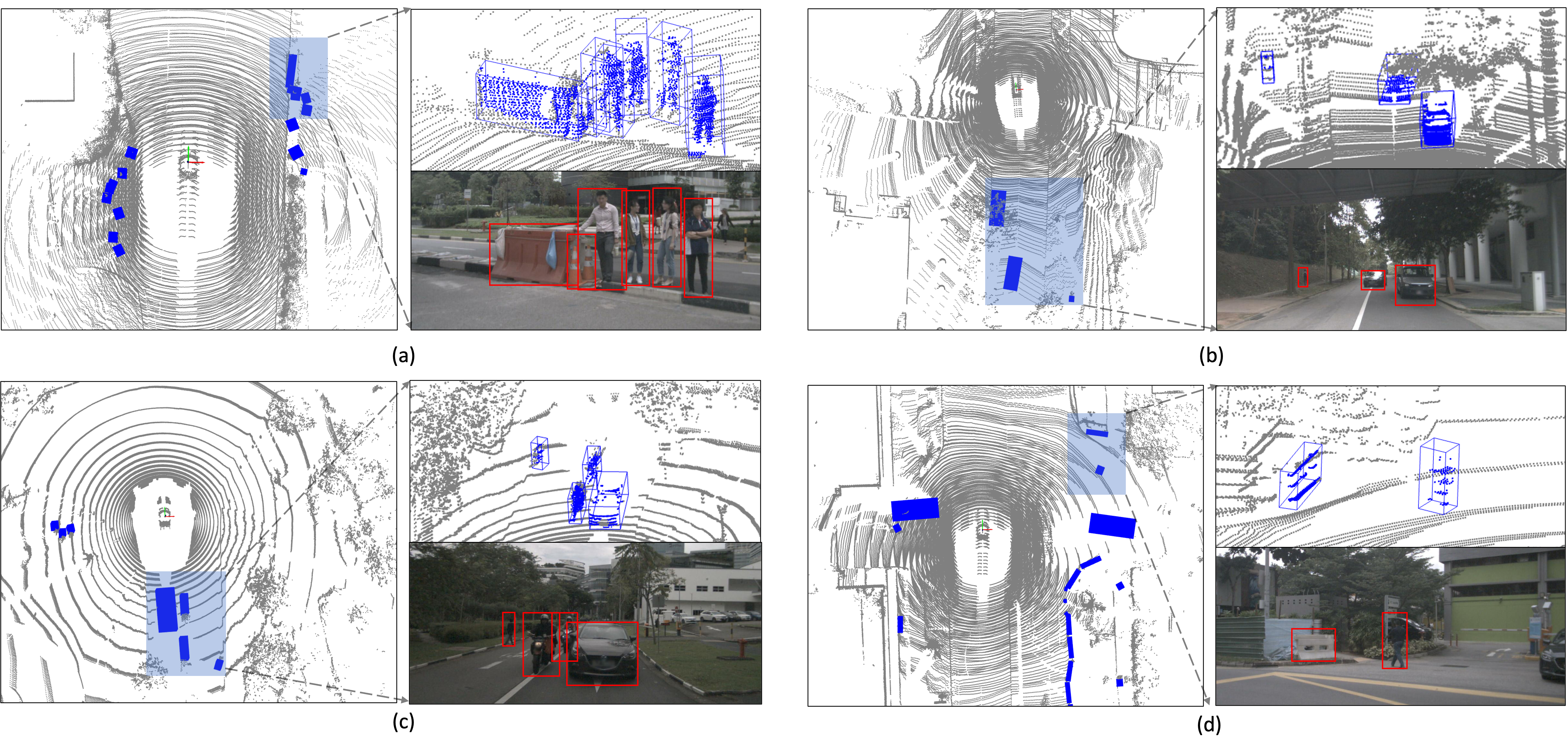}
  \vspace{-1em}
  \caption{Visualization of detected novel classes in different LiDAR frames. (a) Detected `barrier', `traffic\_cone', and `pedestrian'; (b) Detected `car' and `pedestrian'; (c) Detected `car' and `motorcycle'; (d) Detected `barrier' and `pedestrian'.}
  \label{fig: visualization of detection results.}
  \vspace{-1em}
\end{figure*}

As shown in Figure~\ref{fig: visualization of detection results.}, our framework demonstrates a strong ability to detect novel objects by inputting the interested categories. We visualize the results from both BEV and a specific perspective. For example, our method successfully identifies the `pedestrian', `barrier', and `traffic\_cone' in example (a), and the `car' and `motorcycle' in (c). These results also indicate that our method exhibits superior open-vocabulary detection ability across different scenes.

\begin{table}
\setlength{\tabcolsep}{4pt}
\footnotesize
  \centering
  \begin{tabular}{c|cc|ccc}
    \toprule
    \# & Temporal & Spatial &mAP-near &mAP-midrange &mAP-far\\
    \midrule
    (1) & & &27.1 &24.8 & 22.1  \\
    (2) &\checkmark & &27.4 &25.9 & 23.0\\
    (3) & &\checkmark &27.5 &26.0 & 22.8\\
    (4) &\checkmark &\checkmark &28.0 & 26.5 &23.5\\
    \bottomrule
  \end{tabular}
  \vspace{-1em}
  \caption{Results with temporal and spatial awareness \emph{w.r.t} mAP-near, mAP-medium, mAP-far.}
  \label{tab: component in generic object}
  \vspace{-1em}
\end{table}

\subsection{Ablation}
In this part, we present a comprehensive analysis of OpenSight and examine the effectiveness of each component.

\noindent \textbf{Components in Generic Object Perception.}
We decouple the components in generic object perception and record the results in Table~\ref{tab: component in generic object}. Specifically, both modules improve more for mAP-medium and mAP-far. For example, temporal improves 1.1 AP for mAP-medium while 0.3 AP for mAP-near. Both modules interact positively with each other and further improve the results when jointly introduced.

\noindent \textbf{Components in Semantic Interpretation.}
We analyze the modules of direct alignment, 3D-2D feature integration, and LLMs priors in semantic interpretation and record the results in Table~\ref{tab: component in semantic integration}.
The performance is consistently improved with integration and priors introduced. Compared with LLMs priors, 3D-2D integration brings more benefits, for example, 0.9 \emph{vs.} 0.2 in mAP-medium. 

\begin{table}
\setlength{\tabcolsep}{4pt}
\footnotesize
  \centering
  \begin{tabular}{c|ccc|ccc}
    \toprule
    \# & D.A &I.G. &Priors &mAP-large &mAP-medium &mAP-small \\
    \midrule
    (1) &\checkmark & & &9.4 &31.8 &38.1 \\ 
    (2) &\checkmark &\checkmark & &10.9 &32.7 &38.8 \\
    (3) &\checkmark &\checkmark &\checkmark &11.5 &32.9 &39.0\\
    \bottomrule
  \end{tabular}
   \vspace{-1em}
   \caption{Results with direct alignment (`D.A.'), 3D-2D feature integration (`I.G.'), and LLMs priors.}
  \label{tab: component in semantic integration}
  \vspace{-1em}
\end{table}

\noindent \textbf{Image-level \emph{vs.} Object-level.} 
We also investigate the image-level open-vocabulary model CLIP~\cite{radford2021learning} in semantic interpretation. The results are recorded in Table~\ref{tab: different LVLs}. Compared with our adopted object-level grounding DINO, we notice that object-level alignment shows much better results from either a distance perspective or a size perspective. For example, 28.0 AP of near objects is achieved in object-level alignment, 9 points higher than image-level alignment. We analyze that CLIP requires more accurate bounding box prediction, in comparison with object-level alignment which only requires a nearest box. 

\noindent \textbf{Importance of Bounding Boxes Existence.} 
Existing open-vocabulary methods focus more on the alignment between 3D features and 2D features. However, in LiDAR-based object detection, we emphasize that accurate bounding boxes are essential before alignment. We select some classes and investigate both average precision (AP) and average recall (AR) under different thresholds. The results are recorded in Table~\ref{tab: importance of boxes existence}. Compared with AP, we observe a higher recall value in different threshold, indicating that although we can recall the correct objects, more objects with incorrect sizes are also recalled. Such results motivate that if we can obtain a more precise estimation of objects, the results will be improved further.

\begin{table}
\setlength{\tabcolsep}{3.5pt}
\footnotesize
  \centering
  \begin{tabular}{l|ccc|ccc}
    \toprule
    mAP & -near & -midrange & -far &-large &-medium &-small \\
    \midrule
    CLIP &19.0 &11.9 &8.2 &3.9 &8.2 &18.4 \\ 
    \midrule
    \makecell{Grounding \\DINO} &28.0 &26.5 &23.5 &11.5 &32.9 &39.0\\
    \bottomrule
  \end{tabular}
   \vspace{-1em}
  \caption{Results of image-level and object-level models.}
  \label{tab: different LVLs}
  \vspace{-1em}
\end{table}

\begin{table}
\setlength{\tabcolsep}{4.7pt}
\footnotesize
  \centering
  \begin{tabular}{l|ccc|ccc}
    \toprule
    \#CLASS & AP-0.5 & AP-1 & AP-2 & AR-0.5 & AR-1 & AR-2\\
    \midrule
    Car &4.7 &18.3 &45.9 &40.0 &69.0 &84.4\\ 
    Truck &0.2 &3.0 &13.3 & 13.2 &30.0 &51.1\\
    Pedestrian &38.8 &51.6 &57.1 &58.4 &69.3 &74.2\\
    \bottomrule
  \end{tabular}
  \vspace{-1em}
  \caption{AP and AR of some classes under different threshold.}
  \label{tab: importance of boxes existence}
  \vspace{-1em}
\end{table}



\section{Conclusion}
We present OpenSight, a simple and versatile open-vocabulary framework for LiDAR-Based object detection. In the absence of ground-truth boxes and semantic labels, OpenSight first learns to identify and localize a wide range of generic objects. To boost the ability of generic object perception, we incorporate temporal and spatial awareness techniques that recalibrate and improve the visibility of estimated 3D boxes. We also align LiDAR object features with the 2D features from the grounding DINO for semantic interpretation. The extensive experiments demonstrate the effectiveness and generalization ability of the proposed framework, paving the way for further exploration in LiDAR-based open-vocabulary challenges.

{
    \small
    \bibliographystyle{ieeenat_fullname}
    \bibliography{main}

\begin{thebibliography}{64}
\providecommand{\natexlab}[1]{#1}
\providecommand{\url}[1]{\texttt{#1}}
\expandafter\ifx\csname urlstyle\endcsname\relax
  \providecommand{\doi}[1]{doi: #1}\else
  \providecommand{\doi}{doi: \begingroup \urlstyle{rm}\Url}\fi

\bibitem[Adams and Bischof(1994)]{adams1994seeded}
Rolf Adams and Leanne Bischof.
\newblock Seeded region growing.
\newblock \emph{IEEE Transactions on pattern analysis and machine intelligence}, 16\penalty0 (6):\penalty0 641--647, 1994.

\bibitem[Alliegro et~al.(2022)Alliegro, Cappio~Borlino, and Tommasi]{alliegro20223dos}
Antonio Alliegro, Francesco Cappio~Borlino, and Tatiana Tommasi.
\newblock 3dos: Towards 3d open set learning-benchmarking and understanding semantic novelty detection on point clouds.
\newblock \emph{Advances in Neural Information Processing Systems}, 35:\penalty0 21228--21240, 2022.

\bibitem[Bai et~al.(2022)Bai, Hu, Zhu, Huang, Chen, Fu, and Tai]{bai2022transfusion}
Xuyang Bai, Zeyu Hu, Xinge Zhu, Qingqiu Huang, Yilun Chen, Hongbo Fu, and Chiew-Lan Tai.
\newblock Transfusion: Robust lidar-camera fusion for 3d object detection with transformers.
\newblock In \emph{Proceedings of the IEEE/CVF Conference on Computer Vision and Pattern Recognition}, pages 1090--1099, 2022.

\bibitem[Caesar et~al.(2020)Caesar, Bankiti, Lang, Vora, Liong, Xu, Krishnan, Pan, Baldan, and Beijbom]{nuscenes}
Holger Caesar, Varun Bankiti, Alex~H. Lang, Sourabh Vora, Venice~Erin Liong, Qiang Xu, Anush Krishnan, Yu Pan, Giancarlo Baldan, and Oscar Beijbom.
\newblock nuscenes: A multimodal dataset for autonomous driving.
\newblock In \emph{CVPR}, 2020.

\bibitem[Cao et~al.(2023)Cao, Zeng, Xu, and Xu]{cao2023coda}
Yang Cao, Yihan Zeng, Hang Xu, and Dan Xu.
\newblock Coda: Collaborative novel box discovery and cross-modal alignment for open-vocabulary 3d object detection.
\newblock \emph{arXiv preprint arXiv:2310.02960}, 2023.

\bibitem[Cen et~al.(2021)Cen, Yun, Cai, Wang, and Liu]{cen2021open}
Jun Cen, Peng Yun, Junhao Cai, Michael~Yu Wang, and Ming Liu.
\newblock Open-set 3d object detection.
\newblock In \emph{2021 International Conference on 3D Vision (3DV)}, pages 869--878. IEEE, 2021.

\bibitem[Cen et~al.(2022)Cen, Yun, Zhang, Cai, Luan, Tang, Liu, and Yu~Wang]{cen2022open}
Jun Cen, Peng Yun, Shiwei Zhang, Junhao Cai, Di Luan, Mingqian Tang, Ming Liu, and Michael Yu~Wang.
\newblock Open-world semantic segmentation for lidar point clouds.
\newblock In \emph{Computer Vision--ECCV 2022: 17th European Conference, Tel Aviv, Israel, October 23--27, 2022, Proceedings, Part XXXVIII}, pages 318--334. Springer, 2022.

\bibitem[Cheraghian et~al.(2019{\natexlab{a}})Cheraghian, Rahman, Campbell, and Petersson]{cheraghian2019mitigating}
Ali Cheraghian, Shafin Rahman, Dylan Campbell, and Lars Petersson.
\newblock Mitigating the hubness problem for zero-shot learning of 3d objects.
\newblock \emph{arXiv preprint arXiv:1907.06371}, 2019{\natexlab{a}}.

\bibitem[Cheraghian et~al.(2019{\natexlab{b}})Cheraghian, Rahman, and Petersson]{cheraghian2019zero}
Ali Cheraghian, Shafin Rahman, and Lars Petersson.
\newblock Zero-shot learning of 3d point cloud objects.
\newblock In \emph{2019 16th International Conference on Machine Vision Applications (MVA)}, pages 1--6. IEEE, 2019{\natexlab{b}}.

\bibitem[Cheraghian et~al.(2020)Cheraghian, Rahman, Campbell, and Petersson]{cheraghian2020transductive}
Ali Cheraghian, Shafin Rahman, Dylan Campbell, and Lars Petersson.
\newblock Transductive zero-shot learning for 3d point cloud classification.
\newblock In \emph{Proceedings of the IEEE/CVF winter conference on applications of computer vision}, pages 923--933, 2020.

\bibitem[Cheraghian et~al.(2022)Cheraghian, Rahman, Chowdhury, Campbell, and Petersson]{cheraghian2022zero}
Ali Cheraghian, Shafin Rahman, Townim~F Chowdhury, Dylan Campbell, and Lars Petersson.
\newblock Zero-shot learning on 3d point cloud objects and beyond.
\newblock \emph{International Journal of Computer Vision}, 130\penalty0 (10):\penalty0 2364--2384, 2022.

\bibitem[Contributors(2020)]{mmdet3d2020}
MMDetection3D Contributors.
\newblock {MMDetection3D: OpenMMLab} next-generation platform for general {3D} object detection.
\newblock \url{https://github.com/open-mmlab/mmdetection3d}, 2020.

\bibitem[Devlin et~al.(2018)Devlin, Chang, Lee, and Toutanova]{devlin2018bert}
Jacob Devlin, Ming-Wei Chang, Kenton Lee, and Kristina Toutanova.
\newblock Bert: Pre-training of deep bidirectional transformers for language understanding.
\newblock \emph{arXiv preprint arXiv:1810.04805}, 2018.

\bibitem[Douillard et~al.(2011)Douillard, Underwood, Kuntz, Vlaskine, Quadros, Morton, and Frenkel]{douillard2011segmentation}
Bertrand Douillard, James Underwood, Noah Kuntz, Vsevolod Vlaskine, Alastair Quadros, Peter Morton, and Alon Frenkel.
\newblock On the segmentation of 3d lidar point clouds.
\newblock In \emph{2011 IEEE International Conference on Robotics and Automation}, pages 2798--2805. IEEE, 2011.

\bibitem[Fan et~al.(2021)Fan, Xiong, Wang, Wang, and Zhang]{fan2021rangedet}
Lue Fan, Xuan Xiong, Feng Wang, Naiyan Wang, and Zhaoxiang Zhang.
\newblock Rangedet: In defense of range view for lidar-based 3d object detection.
\newblock In \emph{Proceedings of the IEEE/CVF International Conference on Computer Vision}, pages 2918--2927, 2021.

\bibitem[Fischler and Bolles(1981)]{fischler1981random}
Martin~A Fischler and Robert~C Bolles.
\newblock Random sample consensus: a paradigm for model fitting with applications to image analysis and automated cartography.
\newblock \emph{Communications of the ACM}, 24\penalty0 (6):\penalty0 381--395, 1981.

\bibitem[Geiger et~al.(2012)Geiger, Lenz, and Urtasun]{Geiger2012CVPR}
Andreas Geiger, Philip Lenz, and Raquel Urtasun.
\newblock Are we ready for autonomous driving? the kitti vision benchmark suite.
\newblock In \emph{Conference on Computer Vision and Pattern Recognition (CVPR)}, 2012.

\bibitem[Gu et~al.(2022)Gu, Lin, Kuo, and Cui]{gu2022openvocabulary}
Xiuye Gu, Tsung-Yi Lin, Weicheng Kuo, and Yin Cui.
\newblock Open-vocabulary object detection via vision and language knowledge distillation.
\newblock In \emph{International Conference on Learning Representations}, 2022.

\bibitem[Guo et~al.(2020)Guo, Wang, Hu, Liu, Liu, and Bennamoun]{guo2020deep}
Yulan Guo, Hanyun Wang, Qingyong Hu, Hao Liu, Li Liu, and Mohammed Bennamoun.
\newblock Deep learning for 3d point clouds: A survey.
\newblock \emph{IEEE transactions on pattern analysis and machine intelligence}, 43\penalty0 (12):\penalty0 4338--4364, 2020.

\bibitem[Huang et~al.(2020)Huang, Liu, Chen, and Bai]{huang2020epnet}
Tengteng Huang, Zhe Liu, Xiwu Chen, and Xiang Bai.
\newblock Epnet: Enhancing point features with image semantics for 3d object detection.
\newblock In \emph{Computer Vision--ECCV 2020: 16th European Conference, Glasgow, UK, August 23--28, 2020, Proceedings, Part XV 16}, pages 35--52. Springer, 2020.

\bibitem[Kirillov et~al.(2023)Kirillov, Mintun, Ravi, Mao, Rolland, Gustafson, Xiao, Whitehead, Berg, Lo, Doll{\'a}r, and Girshick]{kirillov2023segany}
Alexander Kirillov, Eric Mintun, Nikhila Ravi, Hanzi Mao, Chloe Rolland, Laura Gustafson, Tete Xiao, Spencer Whitehead, Alexander~C. Berg, Wan-Yen Lo, Piotr Doll{\'a}r, and Ross Girshick.
\newblock Segment anything.
\newblock \emph{arXiv:2304.02643}, 2023.

\bibitem[Lang et~al.(2019)Lang, Vora, Caesar, Zhou, Yang, and Beijbom]{lang2019pointpillars}
Alex~H Lang, Sourabh Vora, Holger Caesar, Lubing Zhou, Jiong Yang, and Oscar Beijbom.
\newblock Pointpillars: Fast encoders for object detection from point clouds.
\newblock In \emph{Proceedings of the IEEE/CVF conference on computer vision and pattern recognition}, pages 12697--12705, 2019.

\bibitem[Li et~al.(2022{\natexlab{a}})Li, Zhang, Zhang, Yang, Li, Zhong, Wang, Yuan, Zhang, Hwang, et~al.]{li2022grounded}
Liunian~Harold Li, Pengchuan Zhang, Haotian Zhang, Jianwei Yang, Chunyuan Li, Yiwu Zhong, Lijuan Wang, Lu Yuan, Lei Zhang, Jenq-Neng Hwang, et~al.
\newblock Grounded language-image pre-training.
\newblock In \emph{Proceedings of the IEEE/CVF Conference on Computer Vision and Pattern Recognition}, pages 10965--10975, 2022{\natexlab{a}}.

\bibitem[Li et~al.(2023)Li, Ma, Hou, Shi, Yang, Liu, Wu, Chen, Li, Qiao, et~al.]{li2023logonet}
Xin Li, Tao Ma, Yuenan Hou, Botian Shi, Yucheng Yang, Youquan Liu, Xingjiao Wu, Qin Chen, Yikang Li, Yu Qiao, et~al.
\newblock Logonet: Towards accurate 3d object detection with local-to-global cross-modal fusion.
\newblock \emph{arXiv preprint arXiv:2303.03595}, 2023.

\bibitem[Li et~al.(2022{\natexlab{b}})Li, Yu, Meng, Caine, Ngiam, Peng, Shen, Lu, Zhou, Le, et~al.]{li2022deepfusion}
Yingwei Li, Adams~Wei Yu, Tianjian Meng, Ben Caine, Jiquan Ngiam, Daiyi Peng, Junyang Shen, Yifeng Lu, Denny Zhou, Quoc~V Le, et~al.
\newblock Deepfusion: Lidar-camera deep fusion for multi-modal 3d object detection.
\newblock In \emph{Proceedings of the IEEE/CVF Conference on Computer Vision and Pattern Recognition}, pages 17182--17191, 2022{\natexlab{b}}.

\bibitem[Li et~al.(2021)Li, Wang, and Wang]{li2021lidar}
Zhichao Li, Feng Wang, and Naiyan Wang.
\newblock Lidar r-cnn: An efficient and universal 3d object detector.
\newblock In \emph{Proceedings of the IEEE/CVF Conference on Computer Vision and Pattern Recognition}, pages 7546--7555, 2021.

\bibitem[Liang et~al.(2018)Liang, Yang, Wang, and Urtasun]{liang2018deep}
Ming Liang, Bin Yang, Shenlong Wang, and Raquel Urtasun.
\newblock Deep continuous fusion for multi-sensor 3d object detection.
\newblock In \emph{Proceedings of the European conference on computer vision (ECCV)}, pages 641--656, 2018.

\bibitem[Liang et~al.(2022)Liang, Xie, Yu, Xia, Lin, Wang, Tang, Wang, and Tang]{liang2022bevfusion}
Tingting Liang, Hongwei Xie, Kaicheng Yu, Zhongyu Xia, Zhiwei Lin, Yongtao Wang, Tao Tang, Bing Wang, and Zhi Tang.
\newblock Bevfusion: A simple and robust lidar-camera fusion framework.
\newblock \emph{arXiv preprint arXiv:2205.13790}, 2022.

\bibitem[Liu et~al.(2021)Liu, Deng, Dong, and Hu]{liu2021language}
Bo Liu, Shuang Deng, Qiulei Dong, and Zhanyi Hu.
\newblock Language-level semantics conditioned 3d point cloud segmentation.
\newblock \emph{arXiv preprint arXiv:2107.00430}, 2021.

\bibitem[Liu et~al.(2023)Liu, Zeng, Ren, Li, Zhang, Yang, Li, Yang, Su, Zhu, et~al.]{liu2023grounding}
Shilong Liu, Zhaoyang Zeng, Tianhe Ren, Feng Li, Hao Zhang, Jie Yang, Chunyuan Li, Jianwei Yang, Hang Su, Jun Zhu, et~al.
\newblock Grounding dino: Marrying dino with grounded pre-training for open-set object detection.
\newblock \emph{arXiv preprint arXiv:2303.05499}, 2023.

\bibitem[Liu et~al.(2022)Liu, Tang, Amini, Yang, Mao, Rus, and Han]{liu2022bevfusion}
Zhijian Liu, Haotian Tang, Alexander Amini, Xinyu Yang, Huizi Mao, Daniela Rus, and Song Han.
\newblock Bevfusion: Multi-task multi-sensor fusion with unified bird's-eye view representation.
\newblock \emph{arXiv preprint arXiv:2205.13542}, 2022.

\bibitem[Lu et~al.(2023)Lu, Xu, Wei, Xie, Tomizuka, Keutzer, and Zhang]{lu2023open}
Yuheng Lu, Chenfeng Xu, Xiaobao Wei, Xiaodong Xie, Masayoshi Tomizuka, Kurt Keutzer, and Shanghang Zhang.
\newblock Open-vocabulary point-cloud object detection without 3d annotation.
\newblock In \emph{Proceedings of the IEEE/CVF Conference on Computer Vision and Pattern Recognition}, pages 1190--1199, 2023.

\bibitem[Michele et~al.(2021)Michele, Boulch, Puy, Bucher, and Marlet]{michele2021generative}
Bj{\"o}rn Michele, Alexandre Boulch, Gilles Puy, Maxime Bucher, and Renaud Marlet.
\newblock Generative zero-shot learning for semantic segmentation of 3d point clouds.
\newblock In \emph{2021 International Conference on 3D Vision (3DV)}, pages 992--1002. IEEE, 2021.

\bibitem[Minderer et~al.(2022)Minderer, Gritsenko, Stone, Neumann, Weissenborn, Dosovitskiy, Mahendran, Arnab, Dehghani, Shen, et~al.]{minderer2022simple}
Matthias Minderer, Alexey Gritsenko, Austin Stone, Maxim Neumann, Dirk Weissenborn, Alexey Dosovitskiy, Aravindh Mahendran, Anurag Arnab, Mostafa Dehghani, Zhuoran Shen, et~al.
\newblock Simple open-vocabulary object detection with vision transformers.
\newblock \emph{arXiv preprint arXiv:2205.06230}, 2022.

\bibitem[Peng et~al.(2022)Peng, Genova, Jiang, Tagliasacchi, Pollefeys, Funkhouser, et~al.]{peng2022openscene}
Songyou Peng, Kyle Genova, Chiyu Jiang, Andrea Tagliasacchi, Marc Pollefeys, Thomas Funkhouser, et~al.
\newblock Openscene: 3d scene understanding with open vocabularies.
\newblock \emph{arXiv preprint arXiv:2211.15654}, 2022.

\bibitem[Qi et~al.(2017{\natexlab{a}})Qi, Su, Mo, and Guibas]{qi2017pointnet}
Charles~R Qi, Hao Su, Kaichun Mo, and Leonidas~J Guibas.
\newblock Pointnet: Deep learning on point sets for 3d classification and segmentation.
\newblock In \emph{Proceedings of the IEEE conference on computer vision and pattern recognition}, pages 652--660, 2017{\natexlab{a}}.

\bibitem[Qi et~al.(2017{\natexlab{b}})Qi, Yi, Su, and Guibas]{qi2017pointnet++}
Charles~Ruizhongtai Qi, Li Yi, Hao Su, and Leonidas~J Guibas.
\newblock Pointnet++: Deep hierarchical feature learning on point sets in a metric space.
\newblock \emph{Advances in neural information processing systems}, 30, 2017{\natexlab{b}}.

\bibitem[Qi et~al.(2018)Qi, Liu, Wu, Su, and Guibas]{qi2018frustum}
Charles~R Qi, Wei Liu, Chenxia Wu, Hao Su, and Leonidas~J Guibas.
\newblock Frustum pointnets for 3d object detection from rgb-d data.
\newblock In \emph{Proceedings of the IEEE conference on computer vision and pattern recognition}, pages 918--927, 2018.

\bibitem[Radford et~al.(2021)Radford, Kim, Hallacy, Ramesh, Goh, Agarwal, Sastry, Askell, Mishkin, Clark, et~al.]{radford2021learning}
Alec Radford, Jong~Wook Kim, Chris Hallacy, Aditya Ramesh, Gabriel Goh, Sandhini Agarwal, Girish Sastry, Amanda Askell, Pamela Mishkin, Jack Clark, et~al.
\newblock Learning transferable visual models from natural language supervision.
\newblock In \emph{International conference on machine learning}, pages 8748--8763. PMLR, 2021.

\bibitem[Ren et~al.(2015)Ren, He, Girshick, and Sun]{ren2015faster}
Shaoqing Ren, Kaiming He, Ross Girshick, and Jian Sun.
\newblock Faster r-cnn: Towards real-time object detection with region proposal networks.
\newblock \emph{Advances in neural information processing systems}, 28, 2015.

\bibitem[Shi et~al.(2019)Shi, Wang, and Li]{shi2019pointrcnn}
Shaoshuai Shi, Xiaogang Wang, and Hongsheng Li.
\newblock Pointrcnn: 3d object proposal generation and detection from point cloud.
\newblock In \emph{Proceedings of the IEEE/CVF conference on computer vision and pattern recognition}, pages 770--779, 2019.

\bibitem[Shi et~al.(2020)Shi, Guo, Jiang, Wang, Shi, Wang, and Li]{shi2020pv}
Shaoshuai Shi, Chaoxu Guo, Li Jiang, Zhe Wang, Jianping Shi, Xiaogang Wang, and Hongsheng Li.
\newblock Pv-rcnn: Point-voxel feature set abstraction for 3d object detection.
\newblock In \emph{Proceedings of the IEEE/CVF Conference on Computer Vision and Pattern Recognition}, pages 10529--10538, 2020.

\bibitem[Simon et~al.(2019)Simon, Amende, Kraus, Honer, Samann, Kaulbersch, Milz, and Michael~Gross]{simon2019complexer}
Martin Simon, Karl Amende, Andrea Kraus, Jens Honer, Timo Samann, Hauke Kaulbersch, Stefan Milz, and Horst Michael~Gross.
\newblock Complexer-yolo: Real-time 3d object detection and tracking on semantic point clouds.
\newblock In \emph{Proceedings of the IEEE/CVF Conference on Computer Vision and Pattern Recognition Workshops}, pages 0--0, 2019.

\bibitem[Sindagi et~al.(2019)Sindagi, Zhou, and Tuzel]{sindagi2019mvx}
Vishwanath~A Sindagi, Yin Zhou, and Oncel Tuzel.
\newblock Mvx-net: Multimodal voxelnet for 3d object detection.
\newblock In \emph{2019 International Conference on Robotics and Automation (ICRA)}, pages 7276--7282. IEEE, 2019.

\bibitem[Sun et~al.(2020)Sun, Kretzschmar, Dotiwalla, Chouard, Patnaik, Tsui, Guo, Zhou, Chai, Caine, et~al.]{sun2020scalability}
Pei Sun, Henrik Kretzschmar, Xerxes Dotiwalla, Aurelien Chouard, Vijaysai Patnaik, Paul Tsui, James Guo, Yin Zhou, Yuning Chai, Benjamin Caine, et~al.
\newblock Scalability in perception for autonomous driving: Waymo open dataset.
\newblock In \emph{Proceedings of the IEEE/CVF conference on computer vision and pattern recognition}, pages 2446--2454, 2020.

\bibitem[Sun et~al.(2021)Sun, Wang, Chai, Elsayed, Bewley, Zhang, Sminchisescu, and Anguelov]{sun2021rsn}
Pei Sun, Weiyue Wang, Yuning Chai, Gamaleldin Elsayed, Alex Bewley, Xiao Zhang, Cristian Sminchisescu, and Dragomir Anguelov.
\newblock Rsn: Range sparse net for efficient, accurate lidar 3d object detection.
\newblock In \emph{Proceedings of the IEEE/CVF Conference on Computer Vision and Pattern Recognition}, pages 5725--5734, 2021.

\bibitem[Wang et~al.(2021)Wang, Ma, Zhu, and Yang]{wang2021pointaugmenting}
Chunwei Wang, Chao Ma, Ming Zhu, and Xiaokang Yang.
\newblock Pointaugmenting: Cross-modal augmentation for 3d object detection.
\newblock In \emph{Proceedings of the IEEE/CVF Conference on Computer Vision and Pattern Recognition}, pages 11794--11803, 2021.

\bibitem[Wang et~al.(2020)Wang, Fathi, Kundu, Ross, Pantofaru, Funkhouser, and Solomon]{wang2020pillar}
Yue Wang, Alireza Fathi, Abhijit Kundu, David~A Ross, Caroline Pantofaru, Tom Funkhouser, and Justin Solomon.
\newblock Pillar-based object detection for autonomous driving.
\newblock In \emph{Computer Vision--ECCV 2020: 16th European Conference, Glasgow, UK, August 23--28, 2020, Proceedings, Part XXII 16}, pages 18--34. Springer, 2020.

\bibitem[Wong et~al.(2020)Wong, Wang, Ren, Liang, and Urtasun]{wong2020identifying}
Kelvin Wong, Shenlong Wang, Mengye Ren, Ming Liang, and Raquel Urtasun.
\newblock Identifying unknown instances for autonomous driving.
\newblock In \emph{Conference on Robot Learning}, pages 384--393. PMLR, 2020.

\bibitem[Yan et~al.(2018)Yan, Mao, and Li]{yan2018second}
Yan Yan, Yuxing Mao, and Bo Li.
\newblock Second: Sparsely embedded convolutional detection.
\newblock \emph{Sensors}, 18\penalty0 (10):\penalty0 3337, 2018.

\bibitem[Yang et~al.(2020)Yang, Sun, Liu, and Jia]{yang20203dssd}
Zetong Yang, Yanan Sun, Shu Liu, and Jiaya Jia.
\newblock 3dssd: Point-based 3d single stage object detector.
\newblock In \emph{Proceedings of the IEEE/CVF conference on computer vision and pattern recognition}, pages 11040--11048, 2020.

\bibitem[Yao et~al.(2022)Yao, Han, Wen, Liang, Xu, Zhang, Li, Xu, and Xu]{yao2022detclip}
Lewei Yao, Jianhua Han, Youpeng Wen, Xiaodan Liang, Dan Xu, Wei Zhang, Zhenguo Li, Chunjing Xu, and Hang Xu.
\newblock Detclip: Dictionary-enriched visual-concept paralleled pre-training for open-world detection.
\newblock \emph{arXiv preprint arXiv:2209.09407}, 2022.

\bibitem[Yin et~al.(2021{\natexlab{a}})Yin, Zhou, and Krahenbuhl]{yin2021center}
Tianwei Yin, Xingyi Zhou, and Philipp Krahenbuhl.
\newblock Center-based 3d object detection and tracking.
\newblock In \emph{Proceedings of the IEEE/CVF conference on computer vision and pattern recognition}, pages 11784--11793, 2021{\natexlab{a}}.

\bibitem[Yin et~al.(2021{\natexlab{b}})Yin, Zhou, and Kr{\"a}henb{\"u}hl]{yin2021multimodal}
Tianwei Yin, Xingyi Zhou, and Philipp Kr{\"a}henb{\"u}hl.
\newblock Multimodal virtual point 3d detection.
\newblock \emph{Advances in Neural Information Processing Systems}, 34:\penalty0 16494--16507, 2021{\natexlab{b}}.

\bibitem[Yoo et~al.(2020)Yoo, Kim, Kim, and Choi]{yoo20203d}
Jin~Hyeok Yoo, Yecheol Kim, Jisong Kim, and Jun~Won Choi.
\newblock 3d-cvf: Generating joint camera and lidar features using cross-view spatial feature fusion for 3d object detection.
\newblock In \emph{Computer Vision--ECCV 2020: 16th European Conference, Glasgow, UK, August 23--28, 2020, Proceedings, Part XXVII 16}, pages 720--736. Springer, 2020.

\bibitem[Yurtsever et~al.(2020)Yurtsever, Lambert, Carballo, and Takeda]{yurtsever2020survey}
Ekim Yurtsever, Jacob Lambert, Alexander Carballo, and Kazuya Takeda.
\newblock A survey of autonomous driving: Common practices and emerging technologies.
\newblock \emph{IEEE access}, 8:\penalty0 58443--58469, 2020.

\bibitem[Zang et~al.(2022)Zang, Li, Zhou, Huang, and Loy]{zang2022open}
Yuhang Zang, Wei Li, Kaiyang Zhou, Chen Huang, and Chen~Change Loy.
\newblock Open-vocabulary detr with conditional matching.
\newblock In \emph{Computer Vision--ECCV 2022: 17th European Conference, Tel Aviv, Israel, October 23--27, 2022, Proceedings, Part IX}, pages 106--122. Springer, 2022.

\bibitem[Zareian et~al.(2021)Zareian, Rosa, Hu, and Chang]{zareian2021open}
Alireza Zareian, Kevin~Dela Rosa, Derek~Hao Hu, and Shih-Fu Chang.
\newblock Open-vocabulary object detection using captions.
\newblock In \emph{Proceedings of the IEEE/CVF Conference on Computer Vision and Pattern Recognition}, pages 14393--14402, 2021.

\bibitem[Zhang et~al.(2022{\natexlab{a}})Zhang, Li, Liu, Zhang, Su, Zhu, Ni, and Shum]{zhang2022dino}
Hao Zhang, Feng Li, Shilong Liu, Lei Zhang, Hang Su, Jun Zhu, Lionel~M Ni, and Heung-Yeung Shum.
\newblock Dino: Detr with improved denoising anchor boxes for end-to-end object detection.
\newblock \emph{arXiv preprint arXiv:2203.03605}, 2022{\natexlab{a}}.

\bibitem[Zhang et~al.(2022{\natexlab{b}})Zhang, Guo, Zhang, Li, Miao, Cui, Qiao, Gao, and Li]{zhang2022pointclip}
Renrui Zhang, Ziyu Guo, Wei Zhang, Kunchang Li, Xupeng Miao, Bin Cui, Yu Qiao, Peng Gao, and Hongsheng Li.
\newblock Pointclip: Point cloud understanding by clip.
\newblock In \emph{Proceedings of the IEEE/CVF Conference on Computer Vision and Pattern Recognition}, pages 8552--8562, 2022{\natexlab{b}}.

\bibitem[Zhou et~al.(2022)Zhou, Girdhar, Joulin, Kr{\"a}henb{\"u}hl, and Misra]{zhou2022detecting}
Xingyi Zhou, Rohit Girdhar, Armand Joulin, Philipp Kr{\"a}henb{\"u}hl, and Ishan Misra.
\newblock Detecting twenty-thousand classes using image-level supervision.
\newblock In \emph{Computer Vision--ECCV 2022: 17th European Conference, Tel Aviv, Israel, October 23--27, 2022, Proceedings, Part IX}, pages 350--368. Springer, 2022.

\bibitem[Zhou and Tuzel(2018)]{zhou2018voxelnet}
Yin Zhou and Oncel Tuzel.
\newblock Voxelnet: End-to-end learning for point cloud based 3d object detection.
\newblock In \emph{Proceedings of the IEEE conference on computer vision and pattern recognition}, pages 4490--4499, 2018.

\bibitem[Zhou et~al.(2020)Zhou, Sun, Zhang, Anguelov, Gao, Ouyang, Guo, Ngiam, and Vasudevan]{zhou2020end}
Yin Zhou, Pei Sun, Yu Zhang, Dragomir Anguelov, Jiyang Gao, Tom Ouyang, James Guo, Jiquan Ngiam, and Vijay Vasudevan.
\newblock End-to-end multi-view fusion for 3d object detection in lidar point clouds.
\newblock In \emph{Conference on Robot Learning}, pages 923--932. PMLR, 2020.

\bibitem[Zhu et~al.(2022)Zhu, Zhang, He, Guo, Zeng, Qin, Zhang, and Gao]{Zhu2022PointCLIPV2}
Xiangyang Zhu, Renrui Zhang, Bowei He, Ziyu Guo, Ziyao Zeng, Zipeng Qin, Shanghang Zhang, and Peng Gao.
\newblock Pointclip v2: Prompting clip and gpt for powerful 3d open-world learning.
\newblock \emph{arXiv preprint arXiv:2211.11682}, 2022.

\end{thebibliography}
}

\clearpage
\setcounter{page}{1}
\maketitlesupplementary

\section{LLMs Priors}
In Section~\ref{subsection: generic object perception}, we use size priors to select those ``precisely'' estimated 3D boxes for spatial awareness. The details of size priors from GPT-4 are presented in Table~\ref{tab: size priors}.
\begin{table}[h!]
\small
\centering
\begin{tabular}{@{}lccc@{}}
\toprule
Object                & Width (m) & Length (m) & Height (m) \\ \midrule
Truck                 & 2.52      & 6.94       & 2.85       \\
Pedestrian            & 0.67      & 0.73       & 1.77       \\
Car                   & 1.96      & 4.63       & 1.74       \\
Traffic Cone          & 0.41      & 0.42       & 1.08       \\
Motorcycle            & 0.77      & 2.11       & 1.46       \\
Construction Vehicle  & 2.82      & 6.56       & 3.20        \\
Trailer               & 2.92      & 12.28      & 3.87       \\
Barrier               & 2.51      & 0.50        & 0.99       \\
Bicycle               & 0.61      & 1.70        & 1.30        \\
Bus                   & 2.95      & 11.19      & 3.49       \\ \bottomrule
\end{tabular}
\caption{Size priors of common objects from LLMs in an autonomous driving scene.}
\label{tab: size priors}
\end{table}

We observe that objects in different classes are usually with different shapes, such information is very useful for deciding the good 3D boxes. It is emphasized that although object names are also provided here, we do not use such information and only rely on the width, length, and height information of these objects in spatial awareness.

\section{Evaluation Metrics}
In Section~\ref{sec: experimental settings}, we define the distance-based mAP metrics and size-based metrics. The details are presented in Table~\ref{tab: distance-based evaluation metrics} and Table~\ref{tab: size-based evaluation metrics}, respectively.
\begin{table}[t]
\small
\setlength{\tabcolsep}{25pt}
\centering
\begin{tabular}{@{}lc@{}}
\toprule
Evaluation metrics   & Range (m) \\ 
\midrule
mAP-near                 & 0 $\sim$ 18       \\
mAP-midrange            & 0 $\sim$ 34       \\
mAP-far               & 0 $\sim$ 54  \\
\bottomrule
\end{tabular}
\caption{Criteria for distance-based evaluation metrics.}
\label{tab: distance-based evaluation metrics}
\end{table}

\begin{table}[t]
\small
\setlength{\tabcolsep}{4pt}
\centering
\begin{tabular}{@{}lc@{}}
\toprule
Metrics   & \#CLASSES \\ 
\midrule
mAP-large                 & Car, Truck, Construction vehicle, Bus, and Trailer  \\
mAP-medium            & Barrier, Bicycle, and Motorcycle  \\
mAP-small               & Pedestrian, Traffic cone  \\
\bottomrule
\end{tabular}
\caption{Criteria for size-based evaluation metrics.}
\label{tab: size-based evaluation metrics}
\end{table}

Specifically, for distance-based evaluation, mAP in the detected range 0$\sim$18, 0$\sim$34, 0$\sim$54 is set as mAP-near, mAP-midrange, and mAP-far, respectively. For size-based evaluation, `Car', `Truck', `Construction vehicle', `Bus', and `Trailer' are grouped as `large'; `Barrier', `Bicycle', and `Motorcycle' as `medium'; and `Traffic cone' and `Pedestrian' as `small', leading to corresponding metrics of mAP-large, mAP-medium, and mAP-small.

\section{Implementation Details}
For the 2D boxes in generic object perception, we actually employ the detection results from grounding DINO for the sake of simplicity. It is important to note that this component can seamlessly be substituted with any other detection models, as elaborated in Section~\ref{subsection: generic object perception}. We emphasize that our use of grounding DINO is to derive class-agnostic 2D boxes and ignore the corresponding semantics. Only the location information is needed in that step. The threshold parameter $\tau$ in Equation~\eqref{equ: select box} is set to 0.2, and the distance parameter $d$ in Equation~\eqref{equ: place object} is specified as 54 meters.

\begin{figure*}[t]
  \centering  \includegraphics[width=0.95\textwidth]{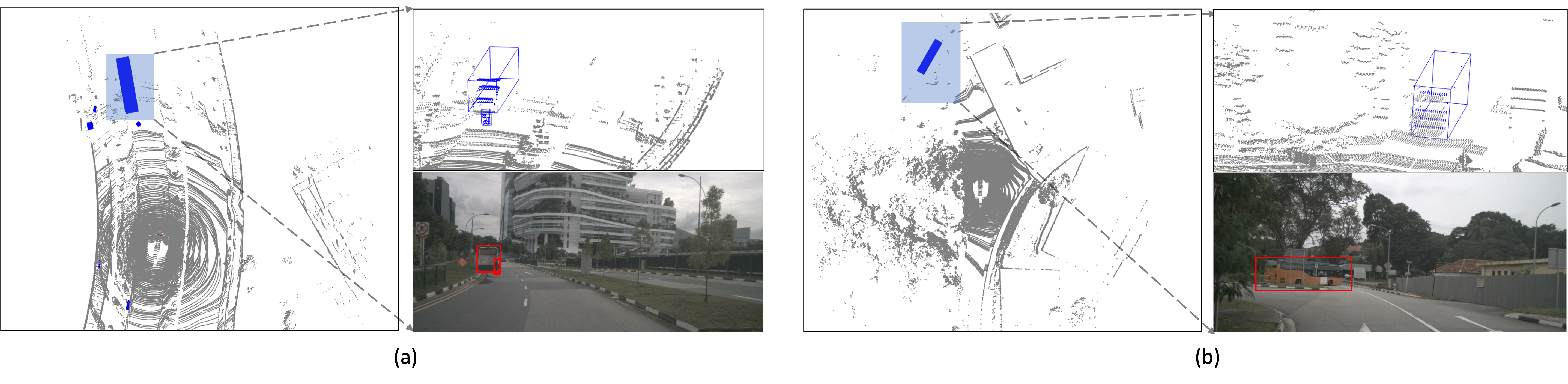}
  \caption{Visualization of detected novel classes in different LiDAR frames. (a) Detected `Construction vehicle' and `pedestrian'; (b) Detected `bus'.}
  \label{fig: visualization of more detection results.}
\end{figure*}
\section{More Discussion}
We analyze more about the obtained results. As illustrated in the main paper, we decouple the whole framework into generic object perception and specific semantic assignment. We consider several cases: (1) The first one is to use both ground-truth 3D boxes and the corresponding labels simultaneously and learn a 3D detection model. It is denoted as `Supervised (box+label)'; (2) The second case uses only ground-truth boxes to learn a class-agnostic detector, ground DINO is then introduced for label assignment. It is denoted as `Supervised (box only)'; (3) The third case also uses only ground-truth boxes to learn a class-agnostic detector, but a fully supervised 2D model is introduced for label assignment (`Supervised 3D box + 2D'). It replaces ground DINO and provides more accurate 2D box features and semantic labels; (4) The fourth case is our generic object detector without ground-truth boxes, but a fully supervised 2D model is introduced for label assignment~(`Our + supervised 2D'); (5) The final case is our generic object detector without ground-truth boxes, and ground DINO is considered for label assignment~(`Our + grounding DINO').
\begin{table*}[t]
\footnotesize
  \centering
  \setlength{\tabcolsep}{6pt}
  \begin{tabular}{l|cc|cccccccccc}
    \toprule
    Method   & mAP & NDS &\scriptsize Car &\scriptsize Truck &\scriptsize C.V. &\scriptsize Bus &\scriptsize Trailer &\scriptsize Barrier &\scriptsize Bicycle &\scriptsize Traffic\_cone &\scriptsize Motorcycle  &\scriptsize Pedestrian\\
    \midrule
    (1)~Supervised~(box+label) & 57.3 &61.7 & 86.0 & 38.8 & 18.3 & 63.3 & 34.2 &70.5 &46.7 &70.2 &59.5 &85.2\\
    (2)~Supervised 3D box + 2D &46.0 &45.5 & 58.0 & 40.4 & 19.0 & 36.4 & 13.8 & 58.3& 49.7 & 60.6 & 54.2 & 69.9\\
    (3)~Supervised~(box only) & 34.2 &39.0 & 53.2 & 20.1 & 7.4 & 20.5 & 0.7 &50.1 &37.8 &52.6 &39.1 &60.8\\
    (4)~Our + supervised 2D &28.3 &25.5 &36.1 &18.3 & 13.3 & 7.0 & 3.0 & 49.4 & 32.1 & 28.2 & 37.4 & 58.8\\
    (5)~Our + grounding DINO & 23.5 &24.0 & 32.1 & 11.6& 8.7 & 5.1 & 0.8 & 42.2 & 26.1 & 25.6 & 30.4 & 52.5\\
    \bottomrule
  \end{tabular}
  \caption{Results on NuScenes dataset. All classes are reserved as novel classes. AP of each class and mAP across all classes are shown. ``Car'' in CoDA is the known class and is fully supervised. (`C.V' is denoted as `construction\_vehicle' here.)}
  \label{tab: map-midrange results}
\end{table*}

By comparing (5) and (4), (3) and (2), we conclude that the open ability of 2D models matters. If we are able to get more powerful 2D models in the open classes, the ability of open detection ability in 3D can be further improved. For example, with a better 2D model, the overall mAP is improved from 23.5\% to 28.3\%.

The comparison between (5) and (3), (4) and (2) also indicates the importance of generic object perception. If we are able to generate more precise 3D bounding boxes for generic objects, the 3D open-vocabulary ability can also be further improved.

We also observe that the performance on `Trailer', `Bus', and `Construction vehicle' are not satisfying. The reasons can be attributed to three aspects. The first reason is that these objects are very large, we are more likely to get an incomplete picture of them in camera images. In LiDAR, we are also likely to get the partial points of these objects. Both of them will increase the difficulty in accurate 3D box estimation. The second reason is that the data is imbalanced and samples of these classes are not sufficient. For supervised learning, we can perform resampling based on the labels, however, for open-vocabulary detection here, label-based resampling is impossible. The final reason is that these classes are fine-grained categories in vehicles, it is hard for ground DINO to detect them accurately.
\section{More Visualization}
We show more visualizations here. For example, our method successfully identifies the `Construction vehicle' in Figure~\ref{fig: visualization of more detection results.} (a) and the `Bus' in Figure~\ref{fig: visualization of more detection results.} (b). These results further indicate the open-vocabulary detection ability of our method in large objects.


\end{document}